\documentclass[postprint,12pt]{elsarticle}%

\usepackage{natbib}
\usepackage[utf8]{inputenc}
\usepackage{graphicx}
\usepackage{amsmath}
\usepackage{amsfonts}
\usepackage{amssymb}
\usepackage{multirow}
\usepackage{array}
\usepackage{booktabs}
\usepackage{cancel}
\usepackage[pdftex]{hyperref}
\usepackage{amsthm}
\usepackage{multicol}
\usepackage{enumerate} 
\usepackage{color}
\hypersetup{colorlinks=true,allcolors=black} 
\usepackage[dvipsnames,svgnames,x11names,hyperref,table]{xcolor}
\usepackage{epsfig}
\usepackage{array}
\usepackage{colortbl}
\usepackage{float}
\usepackage{lineno}
\makeatletter
\def\ps@pprintTitle{%
 \let\@oddhead\@empty
 \let\@evenhead\@empty
 \def\@oddfoot{\footnotesize\itshape
       Postprint submitted to \ifx\@journal\@empty Elsevier
       \else\@journal\fi\hfill}
 \let\@evenfoot\@oddfoot}
\makeatother

\journal{Information Sciences}

\begin{document}

\begin{frontmatter}



\title{Target inductive methods for zero-shot regression}

\author[arcelor]{Miriam Fdez-Díaz \corref{cor1}}
\ead{miriam.fernandezdiaz@arcelormittal.com}
\author[uniovi]{Jos\'e Ram\'on Quevedo}
\ead{quevedo@uniovi.es}
\author[uniovi]{Elena Monta\~n\'es}
\ead{montaneselena@uniovi.es}

\cortext[cor1]{Corresponding author}

\address[arcelor]{ArcelorMittal (Spain)}
\address[uniovi]{Artificial Intelligence Center. University of Oviedo at Gij\'on, 33204 Asturias,  Spain \texttt{http://www.aic.uniovi.es}}

\begin{abstract}
This research arises from the need to predict the amount of air pollutants in meteorological stations. Air pollution depends on the location of the stations (weather conditions and activities in the surroundings). Frequently, the surrounding information is not considered in the learning process. This information is known beforehand in the absence of unobserved weather conditions and remains constant for the same station. Considering the surrounding information as side information facilitates the generalization for predicting pollutants in new stations, leading to a zero-shot regression scenario. Available methods in zero-shot typically lean towards classification, and are not easily extensible to regression. This paper proposes two zero-shot methods for regression. The first method is a similarity based approach that learns models from features and aggregates them using side information. However, potential knowledge of the feature models may be lost in the aggregation. The second method overcomes this drawback by replacing the aggregation procedure and learning the correspondence between side information and feature-induced models, instead. Both proposals are compared with a baseline procedure using artificial datasets, UCI repository communities and crime datasets, and the pollutants. Both approaches outperform the baseline method, but the parameter learning approach manifests its superiority over the similarity based method.
\end{abstract}
\end{frontmatter}
\newpage
\section{Introduction} \label{intro}
The research presented in this paper arises from the desire to predict air pollutants in meteorological stations. Air pollution depends on the location of the stations. On the one hand, weather conditions, which in turn depend on the climate and on the surrounding orography (mountains, sea, etc.), have a notable influence on air pollution.  On the other hand, information about the stations, such as the activities that take place in the surrounding areas (industry, power plants, leisure centers, residential areas, shopping centers, administrative buildings...) also condition air pollution.  Typically, the surrounding information is used merely to split the original task into as many separate tasks as the number of available meteorological stations, and only the weather conditions are considered as features in the learning process \cite{masmoudi2020machine}. Two studies  \cite{delavar2019novel, murillo2019forecasting} represent a preliminary advance in incorporating information about the stations.  The former \cite{delavar2019novel} takes pollutant values from the nearest neighbor station of the target stations, whereas the latter \cite{murillo2019forecasting} performs a prior classification of the stations, taking into account information about roads, traffic flow, or whether they are located in urban, suburban or industrial areas. However, this valuable information about the stations is not included in the learning procedure. This information is known beforehand in the absence of unobserved weather conditions and remains constant for the same meteorological station. This fact allows us to regard this information as being what we usually refer to as side (or privileged) information \cite{vapnik2009}. Side information is neither features nor targets; it actually constitutes additional and, prior information about how targets (a pollutant measure in stations) are related to features (weather conditions).  It may well be a promising information source susceptible to being exploited to improve the accuracy of the predictions. Furthermore, treating information about stations as side information allows us to make predictions over potential future locations of meteorological stations, for which only side information is available, enabling us to state the problem as a zero-shot regression learning task. 

There already exist many approaches to handling zero-shot learning tasks \cite{liao2018semantic, luo2017zero, shigeto2015ridge, yang2015zeroshot}. In fact, researchers typically associate zero-shot learning to computer vision or image classification \cite{luo2017zero, yang2015zeroshot}. Additionally, the existing zero-shot classification methods are specifically designed for classification, usually for image classification. Despite this, there are few studies that cope with other classification applications \cite{shigeto2015ridge}. The main drawback of these existing classification methods is that their strategies are built on the basis of a classification problem and are therefore not extensible to regressions, or if they are, it is a non-trivial and a non-straightforward task \cite{reis2018hyperprocess}. Certainly, we can commonly find regression-based zero-shot methods for zero-shot learning \cite{liao2018semantic, luo2017zero, palatucci2009zero, shigeto2015ridge, yang2015zeroshot}. But we must distinguish these methods, that are specifically designed for the classification zero-shot task, from methods for the actual regression zero-shot task. The fact is that these frequently labelled regression-based zero-shot methods propose regression to learn a projection function in certain phases of the zero-shot classification task, typically in order to map target instances into either feature spaces or side information (semantic) spaces, or even into other kinds of spaces. Hence, they are not applicable to the regression zero-shot task. At present, only two methods able to cope with zero-shot regression tasks exist in the literature \cite{reis2018hyperprocess, zhang2020cazsl}. The first \cite{reis2018hyperprocess} is a preliminary study, whose experiments include a single toy example based on a beta distribution with just one feature and two side information features. In fact, the available software reports the message "expected 1D vector for $x$" when attempting to perform experiments over datasets with higher dimensions in features and side information. The second \cite{zhang2020cazsl} is a deep learning procedure specifically built to predict the future position of a piece that is pushed by a robotic arm, given the present location. The features (which are information from the images of the piece in the present) are taken together with the side information (additional characteristics of the piece). The main drawback of this method is that the strategy to include side information in the zero-shot learning can only be used in neural networks, and not for a general purpose regressor.

The contribution of this paper, then, is to provide methods to cope with a generic purpose regression zero-shot task; that is to say, the strategies will not depend on the regressor. Two new approaches are proposed. The first one learns a model per observed target (station) from features (weather conditions). Then, it takes the side information (surrounding information about the stations) of both observed (stations) and unobserved (possible future locations of stations) targets through a similarity based relationship method that aggregates the feature models predictions.  However, this aggregation strategy has limited generalization power, since it only interpolates feature model predictions that will be bounded within a certain range, discarding direct knowledge from the models themselves, learned from features. In this sense, the second approach constitutes an alternative that transfers the induced feature models knowledge to another learning procedure which is also fed by target side information. Specifically, it is a parameter learning model correspondence method that learns the parameters of the unobserved target models. So far, and to the best of our knowledge (i) none of the existing studies take surrounding information for the stations as side information for predicting air pollution, nor do they generalize air pollution prediction to new unobserved stations. Furthermore,  (ii) none of the studies include feature model knowledge together with side information into a zero-shot learning procedure for regression tasks.

The rest of the paper is organized as follows: Section \ref{related} discusses existing studies related to the research described in this paper. Section \ref{statement} deals with the notation and the problem statement in a zero-shot framework with side information. Following that, the two approaches we propose for zero-shot regression are detailed in Section \ref{proposal}. The experiments carried out and their results are discussed in Section \ref{experiments}. Finally, in Section \ref{conclusions} we draw some conclusions and suggest certain lines of research for future work.

\section{Related Work} \label{related}
Side information is potential information that relates features and targets and is susceptible to being exploited in order to improve the prediction performance. Sometimes this additional information is not available. If that is the case, the common practice in traditional machine learning approaches is usually to discard it, or if it is not discarded, to consider it as common features. The importance of this privileged information is such that in some studies efforts are directed to extracting, discovering or learning it even before carrying out the learning procedure \cite{farias2019learning, qiam2017alternative}. On other occasions, the information is already available because experts have collected it \cite{palatucci2009zero, qiao2016less}. Side information can be available in different forms for features \cite{mollaysa2019learning} or targets \cite{tamchyna16target}. In addtion, there are several ways of making side information available: in a hierarchy \cite{stefan2019combining}, in a response prediction \cite{menon2011response}, in a structure \cite{jacob2008protein} or as feature representation \cite{kang2017incorporating, palatucci2009zero}. Furthermore, this information source could be heterogeneous and multimodal: for instance, in a recommender system context \cite{aktukmak2019aprobabilistic}. 

Among the possibilities opened up by exploiting side information, zero-shot learning \cite{guo2017synthesizing,romera2015embarrassingly} has become more relevant in recent years. Zero-shot tasks \cite{wang2019asurvey} aim to make predictions for new targets for which no observable data is available. Similar scenarios to zero-shot learning are one-shot learning \cite{feifei2006oneshot} and few-shot learning \cite{peng2019fewshot}, which consist of having respectively one and few observable data available for the new targets. Side information in the new targets becomes particularly relevant in the zero-shot scenario, since it is the only available information that can be exploited in order to yield predictions given the lack of observable data for the new targets \cite{kodirov2015unsupervised}. This fact allows the zero-shot scenario to convert the traditional learning paradigm from transductive to inductive, with respect to the targets \cite{wang2019asurvey}.  Conventional machine learning usually assumes transductive learning with regard to the targets, since the models are induced for specific known and observable targets.  In relation to a zero-shot scenario, which is characterized by the existence of unknown and unobservable targets, we can assume both transductive and inductive learning as far as targets are concerned. The difference between them lies in whether to obtain the models for specific (transductive) or for generic unknown and unobservable (inductive) targets. Most of the studies about zero-shot deal with inductive learning for targets, but there are some that keep the transductive learning property \cite{rahman2019transductive}. The great majority of applications of these scenarios are related to computer vision classification tasks \cite{mettes2017spatial, wang2015zero, xu2017transductive, zhang2016zero}. However, they have also been applied to natural language processing  \cite{johnson2017google, levy2017zero}, mobile and wireless security \cite{robyns2017physical}, or human activity recognition \cite{wang2017zero} classification tasks. 

Zero-shot learning (as well as one-shot and few-shot learning) is closely related to transfer learning. In fact, the absence (or scarcity) of observable data for targets provides transfer learning with an important role in this context \cite{farahani2020concise, zhuang2020comprehensive}. In transfer learning, the observable instances and targets are referred to respectively as source domain and source task, whereas unobservable instances and targets are referred to as target domain and target task. The principle of transfer learning is to transfer the knowledge contained in the source domain and source task in order to solve the target task in the target domain. This means transferring the knowledge from observable instances and targets to unobservable instances and targets. Transfer learning can be split into inductive, transductive or unsupervised transfer learning depending on the different existing situations between the source and target domains, and the tasks \cite{pan2010transferlearning}. In the inductive transfer learning setting, the target task is different from the source task, regardless of whether the source and target domains are identical or not. In the transductive transfer learning setting, the source and target tasks are the same, while the source and target domains are different. Finally,  in the unsupervised transfer learning setting, the target task is different from the source task and the main focus is placed on unsupervised learning tasks. Traditionally \cite{pan2010transferlearning}, instances with their target values (labels in classification) have been required in the target domain for inductive transfer learning, whereas no labeled data in the target domain is available when it comes to transductive transfer learning. But in both cases, a number of instances in the target domain are required. The case studied in this paper can be seen as belonging to the category of inductive transfer learning. In addition, the source and target domains are the same (the features are the same in both domains). However, no instances are available in the target domain, and so classic transfer learning strategies are not applicable to our context. More recently, some studies \cite{campagna2020zero, socher2013zero, yang2019zero} have included transfer learning techniques in a zero-shot framework despite lacking instances in the target domain, but have only done this for classification tasks. Typical strategies followed in these cases involve specific tasks for which they were tailor-made, and, hence they are not suitable for a general purpose task. These strategies are, for instance, finding outliers \cite{socher2013zero}, extracting specific image characteristics \cite{yang2019zero}, or generating synthesized dialog instances \cite{campagna2020zero}. But whatever the case, all of them apply specific strategies that are commonly adopted in classification, but that are not easy, or are even impossible, to extend to regression.

The research addressed in this paper fits into a zero-shot rather than into a one-shot or a few-shot scenario, since new targets (stations) lack observable instances (weather conditions). In addition, our task fits into an inductive rather than into a transductive learning framework for both instances and targets because of the need to produce models for generic unknown and unobservable instances (new weather conditions for any station) and targets (potential stations in new locations). Also, it is a zero-shot regression rather than a classification task, since the targets are pollutant concentrations in stations. Fortunately, side information for targets is available and can be exploited in order to produce more accurate pollutant predictions in the potential locations of new stations. This information is presented in the form of homogeneous feature representation, consisting of all the activities that take place in the surroundings of the stations' new locations. 

\section{Target zero-shot framework for regression} \label{statementmethods}
This section addresses the problem statement and also presents a classification of the strategies to cope with it. 

\subsection{Target inductive versus transductive zero-shot statement for regression} \label{statement}
Let us now formally state the zero-shot task in which side information is available on targets. Let $\mathcal{X}$ be the feature space of instances, $\mathcal{S}$ the feature space of targets (also called the semantic space) and $\mathcal{Y}$ the image space of the predictions. Let us denote $\mathcal{T}^{o}=\{t_{i}^{o}\in \mathcal{S} \textnormal{ : } i=1,\ldots , m_{o}\}$ ($\mathcal{T}^{o}=\{t_{i}^{o}\}_{i=1}^{m_o}$ in what follows) as the set of observed targets and $\mathcal{T}^{u}=\{t_{j}^{u}\in \mathcal{S} \textnormal{ : } j=1,\ldots,m_{u}\}$ ($\mathcal{T}^{u}=\{t_{j}^{u}\}_{j=1}^{m_u}$ in what follows\footnote{Later on and without loss of generality we will assume that $m_u=1$, hence, just one unobserved target is referenced and it will be denoted as $t^u$}) as the set of unobserved targets, such that $\mathcal{T}^{o}\cap\mathcal{T}^{u}=\emptyset$. We will differentiate the image space $\mathcal{Y}^o\subset \mathcal{Y}$ from $\mathcal{Y}^u\subset \mathcal{Y}$ for the observed targets $\mathcal{T}^o$ and unobserved targets $\mathcal{T}^u$. Hence, let us define $\mathcal{D}^{o}=\{(x_{i}^{o},y_{i}^{o})\in  \mathcal{X}\times\mathcal{Y}^{o} \textnormal{ : } i=1,\ldots,n_{o}\}$ as the set of observed instances for observed targets and $\mathcal{D}^{u}=\{ (x_{j}^{u},y_{j}^{u})\in  \mathcal{X}\times\mathcal{Y}^{u} \textnormal{ : } j=1,\ldots,n_{u}\}$ as the unobserved instances for unobserved targets, such that $x_{i}^{o}\neq x_{j}^{u}$ where $i=1,\ldots,n_{o}$ and $j=1,\ldots,n_{u}$, that is, observed and unobserved instances differ from each other. Obviously, in general, $y_{i}^{o}\neq y_{j}^{u}$ where $i=1,\ldots,n_{o}$ and $j=1,\ldots,n_{u}$ since $\mathcal{T}^{o}\cap\mathcal{T}^{u}=\emptyset$. In other words, predictions of observed and unobserved instances also differ from each other. 

The formal statement of zero-shot learning varies depending on whether an inductive or a transductive perspective is adopted (see Section \ref{related}). As observed in Section \ref{related}, this paper will assume an inductive perspective in both instances and targets, since the goal is to build a model for unseen new stations for which weather conditions have not been measured yet. However, we shall formalize the goal of zero-shot learning under both perspectives with regard to targets in order to get a better understanding of the differences. 

Under an inductive scenario, the goal of zero-shot learning is to learn a function $f:\mathcal{X}\times \mathcal{S}\rightarrow \mathcal{Y}^u$ from $\mathcal{D}^{o}$ and $\mathcal{T}^{o}$ able to estimate $\mathcal{D}^{u}_{\mathcal{Y}^u}=\{y_{j}^{u} \in \mathcal{Y}^{u} \textnormal{ : } j=1,\ldots,n_{u}\}$ given a generic $\mathcal{D}^{u}_{\mathcal{X}}=\{x_{j}^{u} \in \mathcal{X} \textnormal{ : } j=1,\ldots,n_{u}\}\subset \mathcal{X}$ and a generic $\mathcal{T}^u\subset S$.

Under a transductive scenario, the goal of zero-shot learning is to learn a function $f^u:\mathcal{X}\rightarrow \mathcal{Y}^u$ from $\mathcal{D}^{o}$, $\mathcal{T}^{o}$ and $\mathcal{T}^{u}$ that can estimate $\mathcal{D}^{u}_{\mathcal{Y}^u}=\{y_{j}^{u} \in \mathcal{Y}^{u} \textnormal{ : } j=1,\ldots,n_{u}\}$ given a generic $\mathcal{D}^{u}_{\mathcal{X}}=\{x_{j}^{u} \in \mathcal{X} \textnormal{ : } j=1,\ldots,n_{u}\}\subset \mathcal{X}$.

\begin{figure}[h]
	\centering
	\includegraphics[width=13.5cm, trim={2cm 6cm 1cm 6cm}, clip]{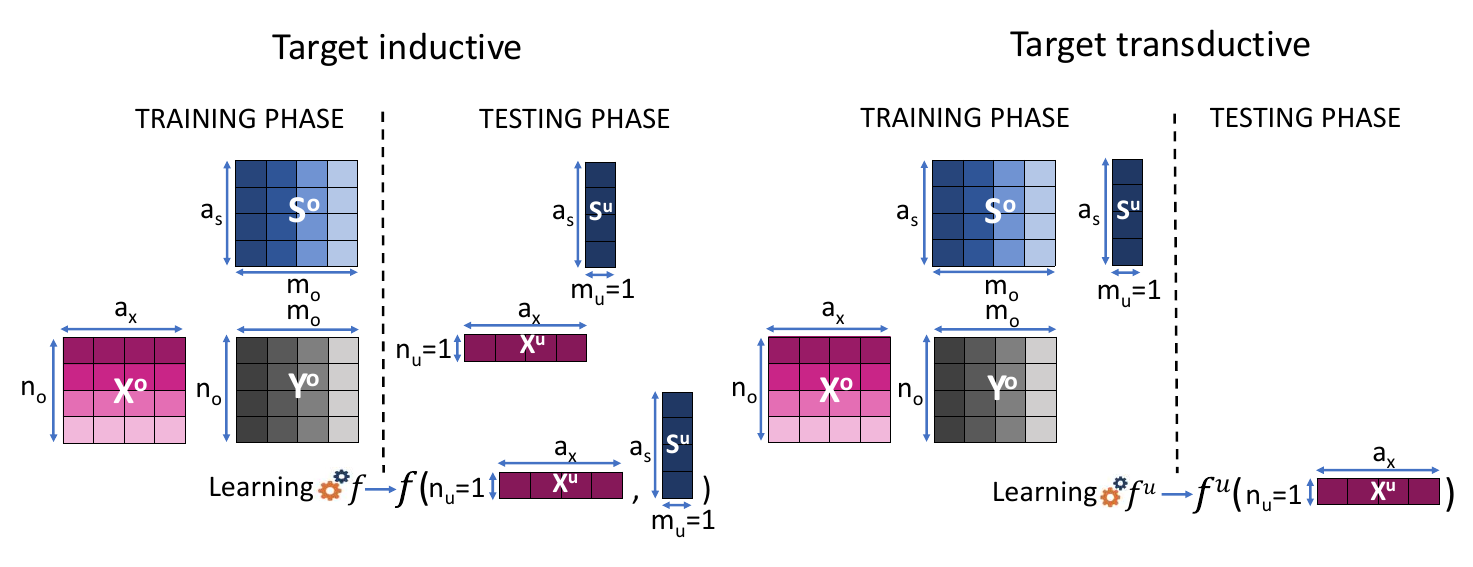}
	\caption{Training and testing phases for both inductive and transductive zero-shot learning using matrix representation.} 
	\label{fig:descriptionTrainTest}
	\centering
\end{figure}

Hence, the difference lies in whether the side information of the unobserved targets $\mathcal{T}^u$ is included in the training phase (transductive) or not (inductive). Figure \ref{fig:descriptionTrainTest} graphically represents such difference using a matrix representation. For this purpose, the $n_{o}$ observed instances of $\mathcal{D}^{o}_{\mathcal{X}}$ and the $n_u$ (equals to $1$ for simplicity) unobserved instances of  $\mathcal{D}^{u}_{\mathcal{X}}$ are respectively represented by the $a_x$ number of features in matrices $X^{o}$ and $X^{u}$ (in purple). Analogously, the $m_o$ observed targets of $\mathcal{T}^o$ and the $m_u$ (equal to $1$ for simplicity) unobserved targets of $\mathcal{T}^u$ are respectively represented by the $a_s$ number of semantic features in matrices $S^o$ and $S^u$ (in blue). Also, the predictions of the $n_o$ observed instances for the $m_o$ observed targets $\mathcal{D}^{o}_{\mathcal{Y}^o}$ are placed in matrix $Y^{o}$ (in grey). Finally, $f$ and $f^u$ predict the $n_u=1$ unobserved instances  of $\mathcal{D}^{u}_{\mathcal{Y}^u}$. In inductive learning, $f$ requires the target unobserved side information $S^{u}$ to make predictions, whereas $f^u$ in transductive learning does not. In this last case, $S^{u}$ is required to learn $f^u$.

Table \ref{tab:notation} summarizes the notation employed.
\begin{table}[H]
	\begin{center}
		\begin{tabular}{cl} 
			\hline
Notation & Explanation\\
			\hline
			$\mathcal{X}$ & feature space of instances\\
			$\mathcal{S}$ & feature space of targets (semantic space)\\
		        $\mathcal{Y}$ & image space of predictions\\
		        $m_o$ & number of observed targets\\
		        $m_u$ & number of unobserved targets\\
		        	$\mathcal{T}^o$ & set of observed targets\\
                         $\mathcal{T}^u$ & set of unobserved targets\\
                         $\mathcal{Y}^o$ & image space of predictions of the observed targets\\
                         $\mathcal{Y}^u$ & image space of predictions of the unobserved targets\\
                         $n_o$ & number of instances of the observed targets\\
		        $n_u$ & number of instances of the unobserved targets\\
                         $\mathcal{D}^o$ & set of instances for the observed targets\\
                         $\mathcal{D}^o_{\mathcal{X}}$ & the projection set of $\mathcal{D}^o$ over the feature space $\mathcal{X}$\\
                         $\mathcal{D}^o_{\mathcal{Y}^o}$ & the projection set of $\mathcal{D}^o$ over the subset $\mathcal{Y}^o\subset\mathcal{Y}$\\
                         $\mathcal{D}^u$ & set of instances for the unobserved targets\\
                         $\mathcal{D}^u_{\mathcal{X}}$ & the projection set of $\mathcal{D}^u$ over the feature space $\mathcal{X}$\\
                         $\mathcal{D}^u_{\mathcal{Y}^u}$ & the projection set of $\mathcal{D}^u$ over the subset $\mathcal{Y}^u\subset\mathcal{Y}$\\
                  
                         $a_x$ & number of features that represent the instances\\
                         $a_s$ & number of features that represent the targets\\
                         $X^o$ & feature matrix representation of the $n_o$ observed instances \\ 
                         &of $\mathcal{D}^o_{\mathcal{X}}$\\
                         $X^u$ & feature matrix representation of the $n_u$ unobserved instances \\ 
                         &of $\mathcal{D}^u_{\mathcal{X}}$\\
                         $S^o$ & semantic feature matrix representation of the $m_o$ observed\\
                         &targets of $\mathcal{T}^o$\\
                         $S^u$ & semantic feature matrix representation of the $m_u$ unobserved\\
                         &targets of $\mathcal{T}^u$\\
                         $Y^o$ & prediction matrix for the $n_o$ observed instances of the $m_o$\\
                         &targets $\mathcal{D}^o_{\mathcal{Y}^o}$\\
                         $Y^u$ & prediction matrix for the $n_u$ unobserved instances of the $m_u$\\
                         &targets $\mathcal{D}^u_{\mathcal{Y}^u}$\\
                         $f$ & the induced function for inductive zero-shot learning\\
                         $f^u$ & the induced function for transductive zero-shot learning\\
                         $f^o$ & the set of the $m_o$ observed targets models\\
		        \hline
		\end{tabular}
		\caption{Summary of the notation (The $Y^u$ matrix is available to test the quality of the predictions, but it is not used in the notation of the paper)}
		\label{tab:notation}
	\end{center}
\end{table}

\subsection{Strategy classification for target zero-shot methods for regression} \label{methods}
Existing zero-shot learning approaches can be split into instance-based methods and classifier-based methods \cite{wang2019asurvey}. However, before continuing, we shall rename the classifier-based methods to model-based methods in order to also include regression methods. Instance-based methods stand out for obtaining (by extracting them, by learning them, and so on) observed instances for the unobserved targets, and afterwards a model for the unobserved targets is learned from these new instances. Model-based methods learn a model for the unobserved targets directly from the information available. 

Let us focus on model-based methods, since all the proposed approaches in this paper aim to obtain a model directly for the unobserved targets. Model-based methods can be split into different types, depending on the strategies they follow to build a model. Specifically, model-based methods fall into the category of relationship methods, correspondence methods and combination methods \cite{wang2019asurvey}: 

\begin{enumerate}
\item [i)] The relationship methods build a model for the unobserved targets from the feature models obtained for the observed targets $f^o=\{f^{t^o_i}\}_{i=1}^{m_o}$ and from a relationship function $\delta^{o,u}$ between observed and unobserved targets. Therefore, the function $f:\mathcal{X}\times \mathcal{S}\rightarrow \mathcal{Y}^u$ learned for the unobserved targets will be a function of $f^o$ and $\delta^{o,u}$. In short, $f(\cdot)=\mathcal{F}(f^o, \delta^{o,u})$.
\item [ii)] The correspondence methods build a model for the unobserved targets learning the correspondence between the side information of the observed targets $\mathcal{T}^o=\{t_i^o\}_{i=1}^{m_o}$ and the feature models obtained for the observed targets $f^o=\{f^{t^o_i}\}_{i=1}^{m_o}$. Therefore, the function $f:\mathcal{X}\times \mathcal{S}\rightarrow \mathcal{Y}^u$ learned for the unobserved targets will be a function of $f^o$ and $\mathcal{T}^o$. That is, $f(\cdot)=\mathcal{F}(f^o, \mathcal{T}^o)$.
\item [iii)] The combination methods are based on decomposing the observed and unobserved targets into basic elements. As a result, the learning process takes place for each of these basic elements. Finally, the models for the basic elements are combined though an inference procedure to get $f:\mathcal{X}\times \mathcal{S}\rightarrow \mathcal{Y}^u$.
\end{enumerate}

The next section will address the methods proposed in this paper.

\section{Target inductive zero-shot methods for regression} \label{proposal}
This paper assumes an inductive perspective in targets (and also in instances) for the zero-shot learning task, as it was detailed in Section \ref{statement}. This approach was motivated by the application that prompted this research, namely, to build a model for unseen new meteorological stations for which weather conditions have not yet been measured. Hence, just the side information from observed targets $\mathcal{T}^o$ is considered in the learning phase.  The side information from unobserved targets $\mathcal{T}^u$ is thus discarded in this phase. The proposed approaches fall into the category of model-based methods, so models for the unobserved targets will be built. One of these approaches classifies as a relationship method, whereas the other belongs to correspondence methods. Hence, although both learn a model per observed target from features $f^o=\{f^{t^o_i}\}_{i=1}^{m_o}$, they differ in how they manage the side information of the observed targets $\mathcal{T}^o$ in order to obtain the models for the unobserved targets. The next subsections detail both of the newly proposed methods and a baseline method for comparison and reference.

\subsection{Baseline} \label{baseline}
\begin{figure}[h]
	\centering
	\includegraphics[width=14.2cm, trim={0 1.9cm 0 1.8cm}, clip]{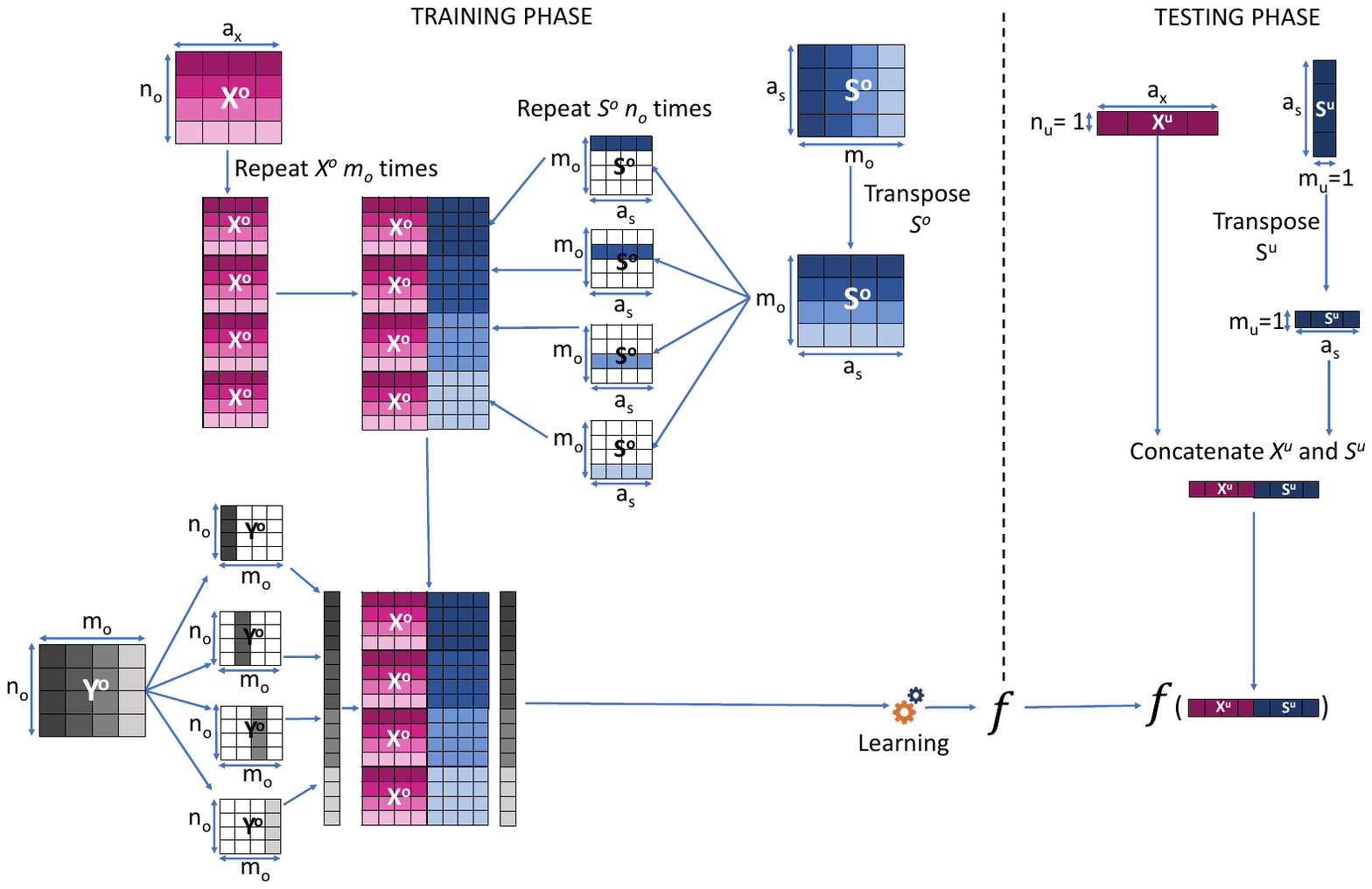}
	\caption{Training and testing phases for the baseline method}
	\label{fig:schemaBaseline}
	\centering
\end{figure}

This section describes a baseline method as a point of reference. The strategy is to treat the side information as common features. Hence, there will be several instances with equal values for the features that correspond to the side information. In the context of air pollution, the instances for weather conditions (features) of the same station will have equal values in the features of the station description (side information). Figure \ref{fig:schemaBaseline} displays the training and testing phases for this baseline approach using the same notation as in Figure \ref{fig:descriptionTrainTest}.

Regarding the training phase (see left side of Figure \ref{fig:schemaBaseline}),  the features for training data for the baseline method are obtained joining features (weather conditions) $X^{o}$ together with side information (station description) $S^{o}$. The target for training data for the baseline method will be the values of $Y^{o}$ (the pollutant concentration given the weather conditions measured in the station description). Therefore, this approach induces just one learning function following a classical machine learning procedure, which corresponds to the function $f:\mathcal{X}\times \mathcal{S}\rightarrow \mathcal{Y}^u$. With regard to the testing phase (see right side of Figure \ref{fig:schemaBaseline}), the features of a new unobserved instance $x^{u}$ (represented by the feature matrix $X^u$) and an unobserved target $t^u$ (represented by the semantic feature matrix $S^{u}$) are joined together to produce the prediction $f(x^u,t^u)$.

\begin{figure}[h]
	\centering
	\includegraphics[width=13.9cm, trim={0 2.5cm 0 2cm}, clip]{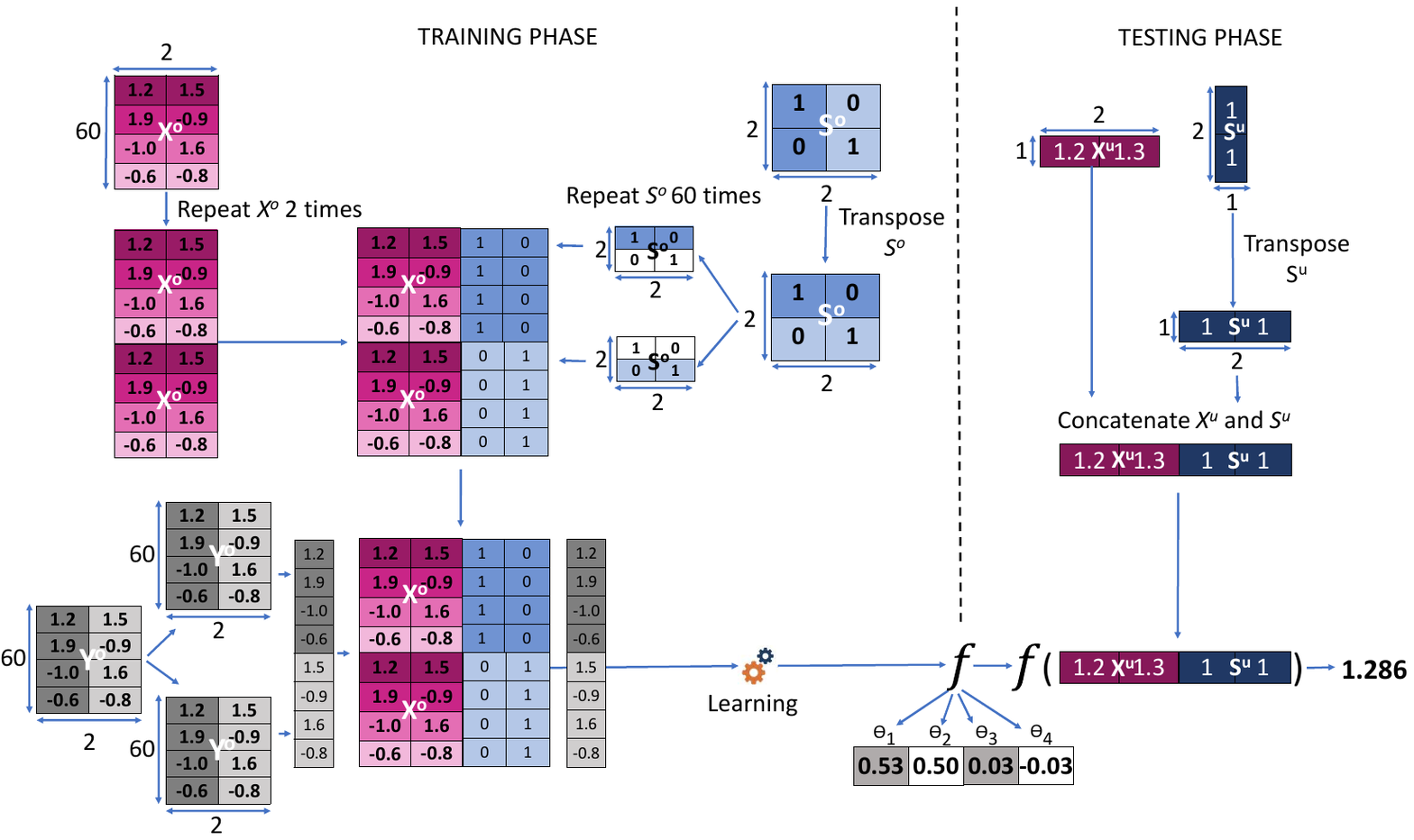}
	\caption{A toy example to illustrate the baseline method}
	\label{fig:exampleExperimentBaseLine}
	\centering
\end{figure}

\subsubsection{A toy example}
Despite the process being simple, let us illustrate it through a toy example with two observed targets, where the function $f$ must estimate $y=x\cdot s$. Figure \ref{fig:exampleExperimentBaseLine} displays the procedure with specific values for $X^{o}$, $S^{o}$, $Y^o$, $X^u$ and $S^u$.  The function $f$ obtained during learning (see parameters $\{\theta_i\}_{i=1}^4$ in the figure) gives low importance to the side information (note that $\theta_3$ and $\theta_4$ are near to zero) and practically averages the features (note that $\theta_1$ and $\theta_2$ are near $0.5$). The prediction of $y$ obtained for the testing instance $(1.2, 1.3)$ is $1.28$, which clearly differs from the actual value of $y=x\cdot s$ which is $2.5$. 
\subsubsection{Discussion}
The main drawback of this baseline method is that different kinds of information are mixed together, assuming the same kind of relationship between features and targets, as between side information and targets. Despite having the same kind of relationship, what actually happens is that features are related to targets through the side information, which means that the side information intervenes in the relationship between features and targets, because of which it does not seem adequate to give equal treatment to both kinds of information. 

\subsection{A relationship zero-shot method for regression} \label{relationship}
\begin{figure}[h]
	\centering
	\includegraphics[width=14cm, trim={1.6cm 2.5cm 1.2cm 1cm}, clip]{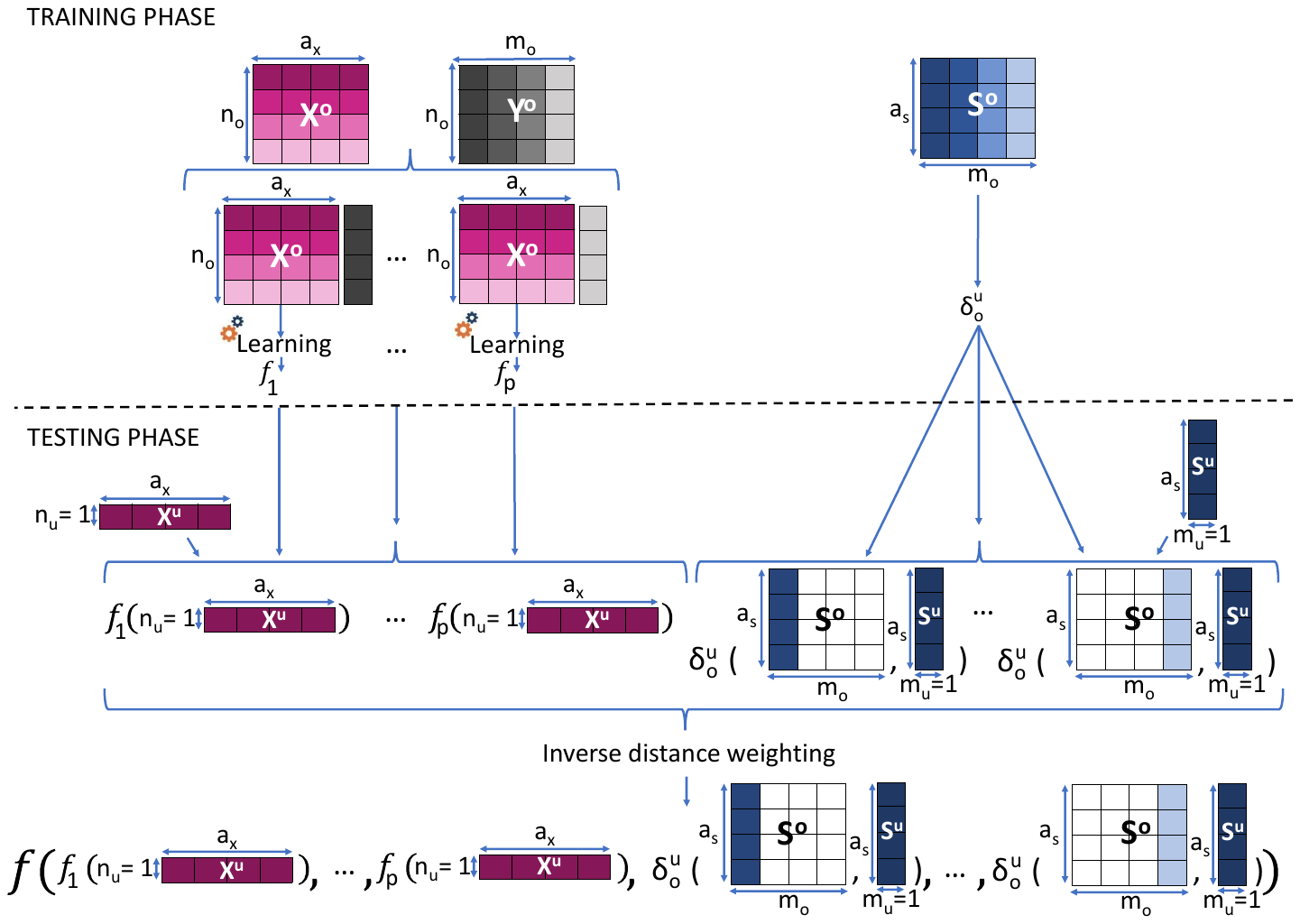}
	\caption{Training and testing phases for the similarity based relationship zero-shot method for regression}
	\label{fig:schemaSINN}
	\centering
\end{figure}

This proposal arises from the necessity to exploit side information in a more adequate way than just giving it the same treatment as the common features. The relationship methods assume a relationship $\delta^{o,u}$ between observed and unobserved targets (see Section \ref{methods}). Then, a function $f:\mathcal{X}\times \mathcal{S}\rightarrow \mathcal{Y}^u$ is induced from $\delta^{o,u}$ and from the models $f^o=\{f^{t^o_i}\}_{i=1}^{m_o}$ of observed targets. Consequently, the proposal consists in inducing the function $f$ in two steps. In the first step, the approach learns the models $f^o=\{f^{t^o_i}\}_{i=1}^{m_o}$ of observed targets solely from the features (weather conditions), thus ignoring the side information (surroundings of the stations) (see the top left side of Figure \ref{fig:schemaSINN}). In the second step, a $\delta^{o,u}$ function is designed in order to perform an aggregation procedure (see the top right side of Figure \ref{fig:schemaSINN}). 
\begin{table}[H]
	\begin{center}
		\begin{tabular}{cl} 
			\hline
Notation & Explanation\\
			\hline
                         $\delta^{o,u}$ & relationship function between observed and unobserved targets\\
                         $\mathcal{T}_k^{t^u}$ & the subset of the closest $k$ observed targets of $\mathcal{T}^o$ to the \\
                         & unobserved target $t^u$\\
			$f_k^{t^u}$ & the subset of models of $f^o$ for the observed targets of $\mathcal{T}_k^{t^u}$\\
              \hline
		\end{tabular}
		\caption{Summary of the notation for the similarity based relationship zero-shot method for regression}
		\label{tab:notationknn}
	\end{center}
\end{table}

We shall now define the $\delta^{o,u}$ function relationship between observed and unobserved targets and the aggregation procedure. 

\subsubsection{Defining the $\delta^{o,u}$ function relationship between observed and unobserved targets}
Our proposal intends to define the $\delta^{o,u}$ relationship in terms of the distance between observed and unobserved targets. As a result, $\delta^{o,u}$ is defined over $\mathcal{T}^o$ and  $\mathcal{T}^u$ and yields a real value that establishes the closeness or similarity between an observed target and an unobserved target through the inverse of the distance, that is,  
$$ \begin{array}{lcll}
\delta^{o,u}:&\mathcal{T}^o\times \mathcal{T}^u&\rightarrow &\mathbb R\\
&(t_i^o,t_j^u)&\rightarrow &1/d(t_i^o,t_j^u)
\end{array}$$
Notice that $\delta^{o,u}$ is strictly positive, since  $\mathcal{T}^o\cap \mathcal{T}^u=\emptyset$, as stated in Section \ref{statement}. Manhattan (L1 norm) and Euclidean (L2 norm) distances could be some examples of a distance $d$ to define $\delta^{o,u}$, typically adopted in the k-nearest neighbor method. 

\subsubsection{Defining the aggregation procedure}
The aggregation procedure takes place in the testing phase (see the bottom of Figure \ref{fig:schemaSINN}) and it is inspired on strategies followed by the $k$-nearest neighbor and the inverse distance weighting methods. For this reason we will call this approach as similarity based relationship method. Firstly, the $k$-nearest observed targets from $\mathcal{T}^o=\{t_i^o\}$ to an unobserved target $t^u$ according to the relationship $\delta^{o,u}$ are chosen. Let's call this subset of observed targets $\mathcal{T}^{t^u}_k$. Remember that $\delta^{o,u}$ has been defined in terms of the inverse of a distance, so $\delta^{o,u}$ measures the closeness or similarity between observed and unobserved targets. Then, only the models from $f^o=\{f^{t^o_i}\}_{i=1}^{m_o}$ of these $k$-nearest observed targets will be taken into account to compute the prediction of the unobserved target $t^u$. Let call this subset of models $f^{t^u}_k=\{f^{t^o_i}\}_{t^o_i \in \mathcal{T}^{t^u}_k}$. Table \ref{tab:notationknn} summarizes the additional notation employed in the development of the similarity relationship zero-shot method for regression.

Secondly, the aggregation procedure is based on inverse distance weighting using the similarity function $\delta^{o,u}$. For this purpose, the features of the new unobserved instance $x^{u}$ feed the set of models $f^{t^u}_k=\{f^{t^o_i}\}_{t^o_i \in \mathcal{T}^{t^u}_k}$ to get the predictions $\{f^{t^o_i}(x^{u})\}_{t^o_i \in \mathcal{T}^{t^u}_k}$. Then, the similarity $\delta^{o,u}(t_i^o,t^u)$ values between the $k$-nearest observed targets $t^o_i \in \mathcal{T}^{t^u}_k$ and the unobserved target $t^u$ are taken. Finally, our proposal entails weighting these predictions with the correspondent similarity values. Hence, the closer the similarity between an unobserved target $t^u$ and the observed target $t^o_i$, the greater influence this observed target model $f^{t^o_i}$ has on the prediction of $t^u$. Therefore, the $f:\mathcal{X}\times \mathcal{S}\rightarrow \mathcal{Y}^u$ is computed as the normalized weighted (by $\{\delta^{o,u}(t_i^o,t^u)\}_{t^o_i \in \mathcal{T}^{t^u}_k}$) average of the $\{f^{t^o_i}(x^u)\}_{t^o_i \in \mathcal{T}^{t^u}_k}$. That is,

$$f(x^u, t^u)=\frac{1}{\sum_{t^o_i \in \mathcal{T}^{t^u}_k}{\delta^{o,u}(t_i^o,t^u)}}\cdot \sum_{t^o_i \in \mathcal{T}^{t^u}_k}\delta^{o,u}(t_i^o,t^u)\cdot f^{t^o_i}(x^u)$$

Notice that the effect of each neighbor is weighted by the inverse of a distance. As a result, the farthest neighbors will have significantly less effect than the nearest neighbors. This illustrates the reason for taking all observed targets instead of just taking the $k$-nearest neighbours. Effectively, their own weights tend to mitigate the influence of the farthest observed targets, avoiding, in addition, the hassle of tuning the hyperparameter $k$. In this context of zero-shot learning, this practice may be promising, since the number of targets is considerably low in relation to the number of instances where methods like $k$-nearest neighbor are applied. Therefore, the low cardinality of $\mathcal{T}^o$ favors not having so many distant observed targets that may have an influence. Nevertheless, with a high number of targets, one can keep just the $k$ closer observed targets.

\begin{figure}[h]
	\centering
	\includegraphics[width=13cm, trim={2cm 1.5cm 4.5cm 1cm}, clip]{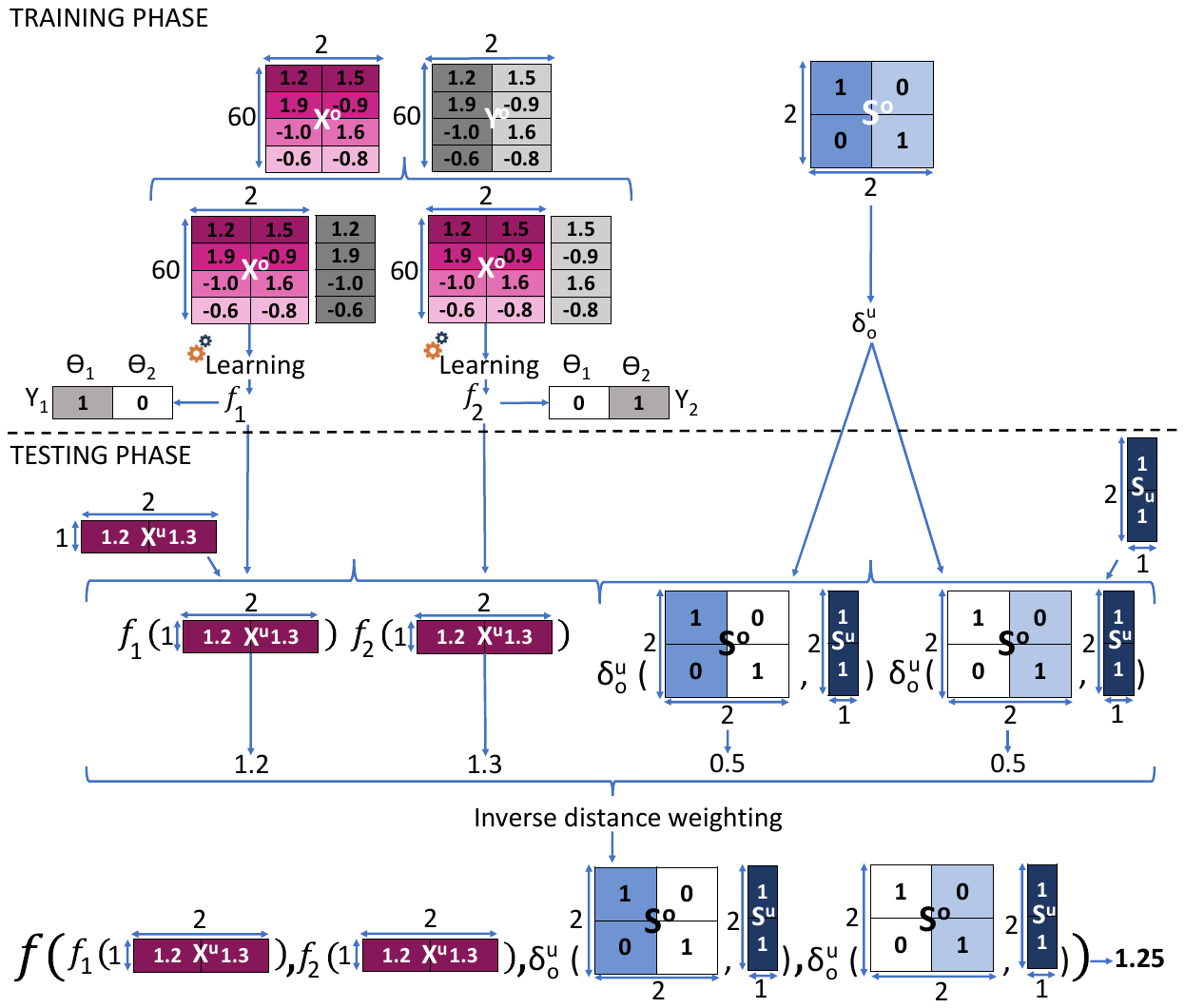}
	\caption{A toy example to illustrate the similarity based relationship zero-shot method for regression}
	\label{fig:exampleExperimentSR}
	\centering
\end{figure}

\subsubsection{Toy example}
Figure \ref{fig:exampleExperimentSR} displays how the process of this method works for the toy example. The expected parameters $\theta$ for the models of the two observed targets are $\theta_{1}=1$ and $\theta_{2}=0$ for $t^o_1$, and $\theta_{1}=0$ and $\theta_{2}=1$ for $t^o_2$, since $y_1=x_1$ and $y_2=x_2$. Once these observed target models are learned, the aggregation procedure takes place. The similarity (inverse of distances) for each observed target is received. In this case, there is no observed target with distance $0$ to the unobserved target, so all the observed targets must be taken into account in order to weight the predictions. The distances obtained from the unobserved target are $0.5$ with regard to both observed targets using the Manhattan distance (L1 norm) (as an example). In order to make predictions for the unobserved target, whose side information is $(1,1)$, the models of the two observed targets are used. The weighted average equals the average, since both similarities are equal to $0.5$ and positive. Consequently, the prediction value $y$ for $x=(1.2,1.3)$ is $1.25$, which also differs from the actual value $y=x\cdot s$, which is $2.5$. 

\subsubsection{Discussion}
The main drawback of this procedure is that the side information of the observed targets $\mathcal{T}^o$, which relates the features with the predictions, is ignored in the learning process, since it is taken into account a posteriori in the testing phase. Hence, knowledge from the feature models $f^o=\{f^{t^o_i}\}_{i=1}^{m_o}$ can be lost. Moreover, it has limited generalization power, since a weighing process just interpolates the predictions produced by $f^o=\{f^{t^o_i}\}_{i=1}^{m_o}$, whose values fall within a certain range. Therefore, the prediction for unobserved targets is limited to this range. Nevertheless, it is an attractive method for dealing with a zero-shot regression task. 

\subsection{A correspondence zero-shot method for regression}\label{correspondence}
This proposal arises from an attempt to overcome the drawbacks of the similarity based relationship method introduced in Section \ref{relationship}. The idea is to include the side information into the learning process in order to increase the generalization power of the predictions for the unobserved targets. The proposal becomes one of the correspondence methods, since it builds the unobserved target models though a learning process in which both the side information of the observed targets $\mathcal{T}^o=\{t_i^o\}_{i=1}^{m_o}$ and the observed target models $f^o=\{f^{t^o_i}\}_{i=1}^{m_o}$ are involved. 

\begin{figure}[h]
	\centering
	\includegraphics[width=13cm, trim={4cm 1.5cm 4cm 1cm}, clip]{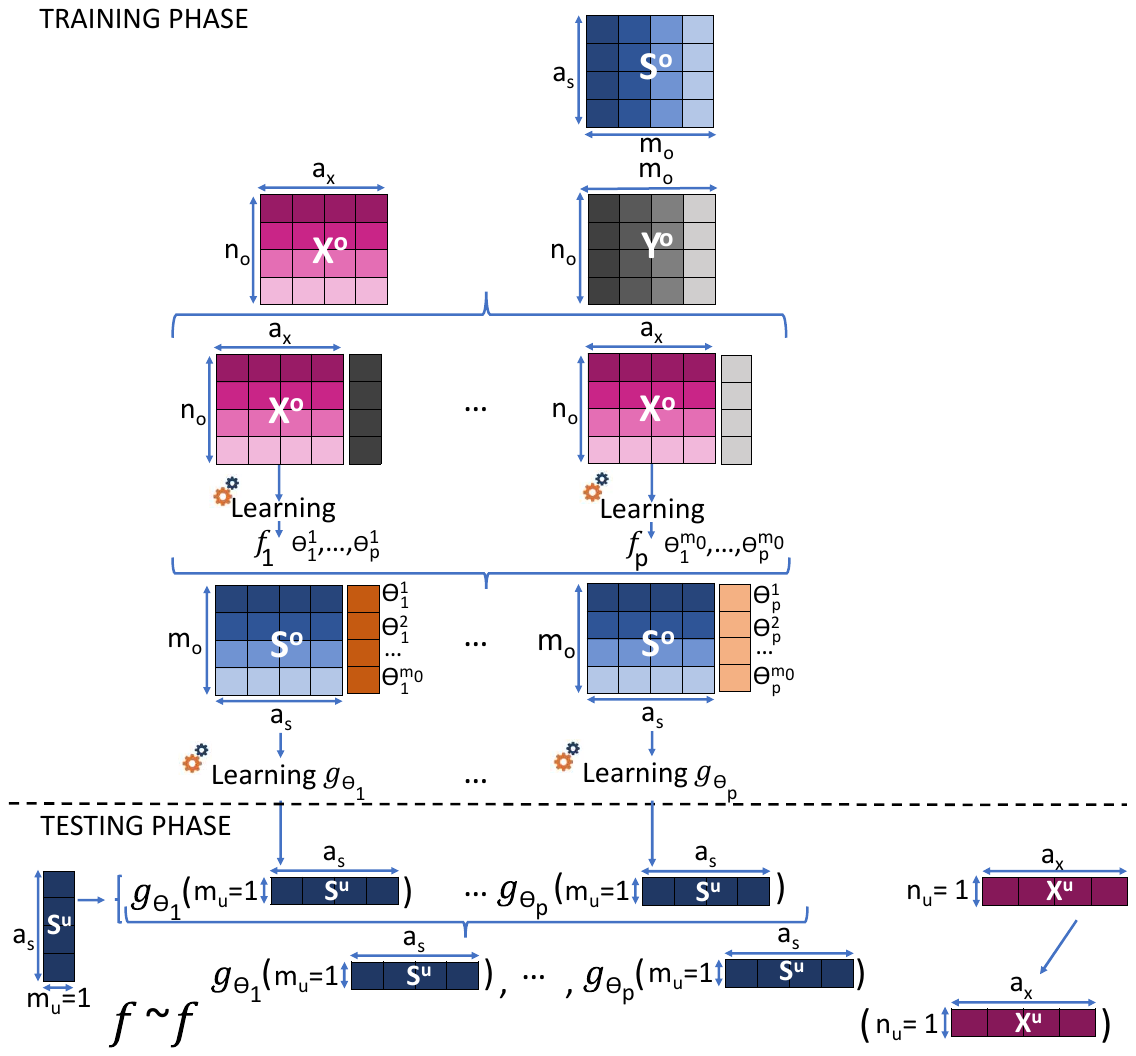}
	\caption{Training and testing phases for the model parameter learning based correspondence zero-shot method for regression}
	\label{fig:schemabSIGenModel}
	\centering
\end{figure}

\begin{figure}[h]
	\centering
	\includegraphics[width=12cm, trim={7cm 5cm 6cm 5.4cm}, clip]{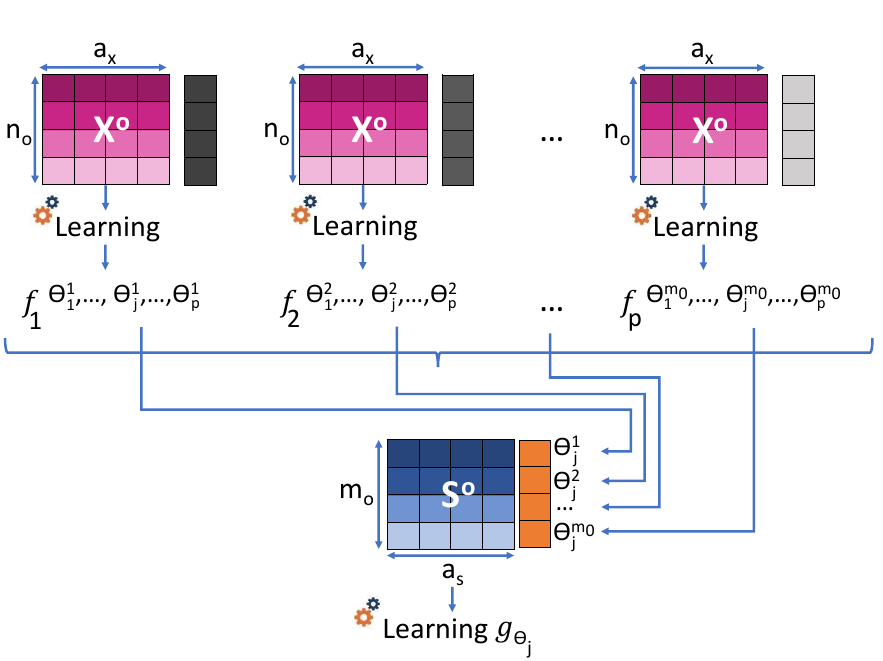}
	\caption{Zoom in learning process of the the model parameter learning based correspondence zero-shot method for regression}
	\label{fig:LearningSIGenModel}
	\centering
\end{figure}
The proposed approach also induces the function $f$ in two steps (see the top of Figure \ref{fig:schemabSIGenModel}). In the first step, the approach learns the models $f^o=\{f^{t^o_i}\}_{i=1}^{m_o}$ of observed targets just from the features, that is, ignoring the side information, much like the similarity based relationship method does (see the top of the training phase of Figure \ref{fig:schemabSIGenModel}). However, in the second step both methods differ. In this new approach, the second step includes a learning process. The idea is to transfer the knowledge of the observed target models $f^o=\{f^{t^o_i}\}_{i=1}^{m_o}$ into the learning process of the unobserved target models. On the one hand, the learned observed target models $f^o=\{f^{t^o_i}\}_{i=1}^{m_o}$ are defined by a set of parameters $\Theta=\{(\theta_1^i,\cdots, \theta_p^i)\}_{i=1}^{m_o}$, which, to some extent, represents the relationship between features and targets. On the other hand, the side information of $\mathcal{T}^o=\{t_i^o\}_{i=1}^{m_o}$ represents the observed targets. In this way, our proposal attempts to learn the correspondence between the side information of the observed targets $\mathcal{T}^o=\{t_i^o\}_{i=1}^{m_o}$ and the feature observed target model parameters $\Theta=\{(\theta_1^i,\cdots, \theta_p^i)\}_{i=1}^{m_o}$. For this reason we will call this approach as model parameter learning based correspondence method. Table \ref{tab:notationmplc} summarizes the additional notation employed in the development of this method.

\begin{table}[H]
	\begin{center}
		\begin{tabular}{cl} 
			\hline
Notation & Explanation\\
			\hline
                  	$p$ & number of parameters for the observed and\\
	& unobserved models\\
			$(\theta_1^i,\cdots, \theta_p^i)$ & the set of $p$ parameters for the model of an observed\\
			& target $i$ of $\mathcal{T}^o$\\
                         $\Theta$ &the set of parameters for the models of all the $m_o$ \\ 
                         &observed targets of $\mathcal{T}^o$\\
                         $g_{\theta_j}$ & induced model for learning the parameter $\theta_j$ of an\\
                         & unobserved target\\
                         $g_{\theta}$ & the set of the $p$ induced models for learning the \\
                         &parameters $(\theta_1,\cdots, \theta_p)$ of an unobserved target\\
		        \hline
		\end{tabular}
		\caption{Summary of the notation for the model parameter learning based correspondence zero-shot method for regression}
		\label{tab:notationmplc}
	\end{center}
\end{table}

Let us detail how this learning process takes place.

\subsubsection{Learning the correspondence between side information and models of observed targets}
This learning process takes place in the second step of the training phase (see the bottom of the training phase of Figure \ref{fig:schemabSIGenModel}). In this step, a set of $p$ learning processes are carried out, as many as the number of parameters $\Theta=\{(\theta_1^i,\cdots, \theta_p^i)\}_{i=1}^{m_o}$ of the observed target models $f^o=\{f^{t^o_i}\}_{i=1}^{m_o}$, which matches the number of parameters for the unobserved target models. Hence, the goal is precisely to learn the parameters of the function $f$ for each of the unobserved targets of $\mathcal{T}^u$. Then, the $g_\theta=\{g_{\theta_j}\}_{j=1}^p$ set of $p$ models is induced, one per parameter of $\{\theta_j\}_{j=1}^p$ for the function $f$ of an unobserved target. The key aspect of these learning processes is the way the datasets constructed to induce the set of models $g_\theta=\{g_{\theta_j}\}_{j=1}^p$ are built. The instances of these datasets are the observed targets $\mathcal{T}^o=\{t_i^o\}_{i=1}^{m_o}$. The features that represent these instances are the side information of the observed targets $\mathcal{T}^o=\{t_i^o\}_{i=1}^{m_o}$. These features are the same for learning all the models $g_\theta=\{g_{\theta_j}\}_{j=1}^p$. Nevertheless, the targets of these instances differ from one model of $g_\theta=\{g_{\theta_j}\}_{j=1}^p$ to another. Specifically, Figure \ref{fig:LearningSIGenModel} details the learning process for inducing $g_{\theta_j}$ that will provide the value of the parameter $\theta_j$ for the function $f$ of an unobserved target $t^u$. The target value of the instance $i$ for $i=1,\dots, m_o$ is the value of the parameter $\theta_i^j$ of the model $f^{t^o_i}$. Hence, the values of the target for the different instances in order to learn the model $g_{\theta_j}$ is the set of values $\{\theta_j^i\}_{i=1}^{m_o}$ of the observed target models $f^o=\{f^{t^o_i}\}_{i=1}^{m_o}$. Therefore, the problem is reduced to a classic regression. From that point on, the general purpose regression algorithm must be applied to learn the set of models $g_\theta=\{g_{\theta_j}\}_{j=1}^p$.

\begin{figure}[p]
	\centering
	\includegraphics[width=16cm, trim={5cm 1cm 2cm 1cm}, clip]{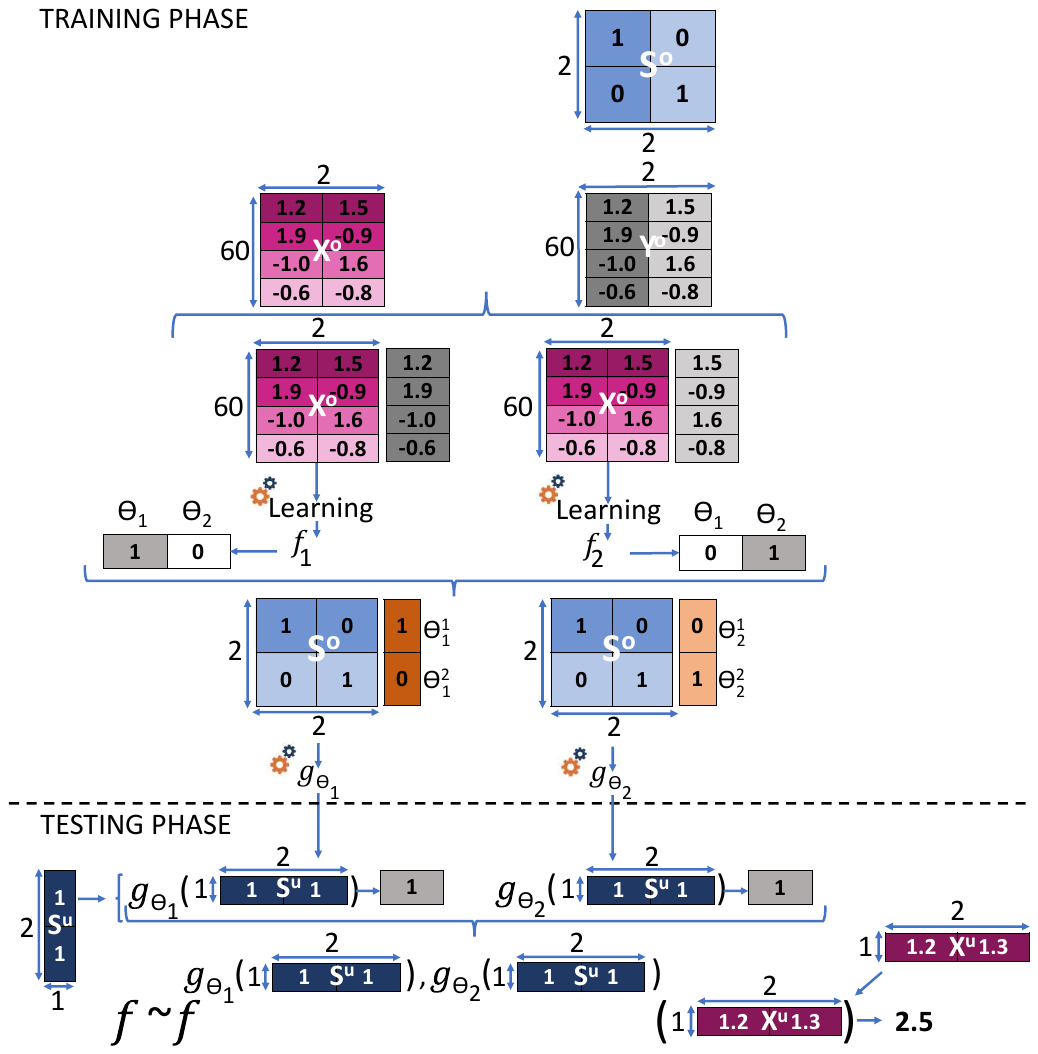}
	\caption{A toy example to illustrate the model parameter learning based correspondence zero-shot method for regression}
	\label{fig:exampleExperimentMPLC}
	\centering
\end{figure}

In the testing phase (see the bottom of Figure \ref{fig:schemabSIGenModel}), the side information of an unobserved target $t^u$ is taken into all the $g_\theta=\{g_{\theta_j}\}_{j=1}^p$ in order to predict the parameters $\{\theta_j^u\}_{j=1}^p$ for the model of the unobserved target $t^u$. Finally, the $f:\mathcal{X}\times \mathcal{S}\rightarrow \mathcal{Y}^u$  is built using these parameter values $\{\theta_j^u=g_{\theta_j}(t^u)\}_{j=1}^p$ and evaluated over an unobserved instance $x^u$. The $g_\theta=\{g_{\theta_j}\}_{j=1}^p$ models will be the correspondence functions between the side information and the feature learned models, whose knowledge will be transferred to each of the unobserved target models. As a result, this method can be considered a model parameter learning based correspondence method, since it learns the parameters for each of the unobserved target models.

\subsubsection{Toy example}
Figure \ref{fig:exampleExperimentMPLC} shows how this method works when applied to the toy example. The models for the two observed targets are the same from the previous method, namely, $\theta_{1}=1$ and $\theta_{2}=0$ for $t^o_1$, and $\theta_{1}=0$ and $\theta_{2}=1$ for $t^o_2$. The models $g_{\theta}$ for the parameters of the unobserved target model are now learned and then applied to the unobserved target. The parameter models are $g_{\theta_{1}}(s)=s_1\cdot1+s_2\cdot0$ because $\theta^{1}_{1}=1$ and $\theta^{2}_{1}=0$, and $g_{\theta_{2}}(s)=s_1\cdot0+s_2\cdot1$ because $\theta^{2}_{1}=0$ and $\theta^{2}_{2}=1$. Hence, the model for the unobserved target whose side information is $(1,1)$ will have $g_{\theta_{1}}((1,1))=1$ and $g_{\theta_{2}}((1,1))=1$ as parameters. Therefore, evaluating the unobserved instance $(1.2,1.3)$ over this model, the prediction obtained will be $2.5$, which is the actual value.

\subsection{Discussion}
As we have shown, the knowledge of the models that is learned from features is transferred together with the side information to the unobserved targets models. This practice increases the generalization prediction power of unobserved instances for unobserved targets, since the model for the unobserved target results from a learning procedure, rather than from an aggregation procedure. 

\section{Experiments} \label{experiments}
This section describes the experiments carried out. Subsection \ref{datasets} describes the datasets taken for the experiments, including the dataset concerning air pollution that motivates this work. In subsection \ref{settings} the parameter settings are established. Finally, subsection \ref{results} analyzes and discusses the results.

\subsection{Description of datasets}\label{datasets}
There are not many datasets in the literature which include side information for regression. The Communities and Crime dataset\footnote{Based on http://archive.ics.uci.edu/ml/datasets/communities+and+crime} of the UCI Machine Learning Repository was suitable for adaptation to this purpose, so it is one of the datasets used for the experiments. In addition, some artificial datasets were created in order to check the performance of the different approaches in a more exhaustive way. Of course, the air pollution dataset was also included. The next subsections provide further details about these different datasets.

\subsubsection{Artificial datasets}
Several artificial datasets were generated in order to study the performance of the methods. The feature values for the instances and the targets were drawn from a uniform distribution in the range of $(-2;-1] \cup [+1;+2)$ in order to avoid zero values or values near zero. This process builds the matrices $X$ and $S$. The $Y$ matrix was carefully built from the matrices $X$ and $S$. 
Hence, if $x=(x_i,\dots,x_{a_x})$ is an instance description and $s=(s_i,\dots,s_{a_s})$ is the side information of a certain target, the value of $y$ for this instance is fixed as a linear function of $x$, where the coefficients are in turn functions of $s$. That is:
\[
y=\sum_{i=1}^{a_x}(\alpha_i(s)\cdot x_i)+\beta
\]

At this point, two kind of artificial datasets are generated, depending on how the $\{\alpha_i(s)\}_{i=1}^{a_x}$ functions are built. In this sense,  the $\{\alpha_i(s)\}_{i=1}^{a_x}$ functions are defined in an attempt to globally cover the domains that both the similarity based relationship method (SR) and the model parameter learning based correspondence method (MPLC) are able to stand up to.
Doing something similar to the baseline method would imply that the side information would not define the relationship between the instance description and the targets. This would mean to dealing with a general purpose regression problem, which is not the aim of this study. Let us now return to the definition of the $\{\alpha_i(s)\}_{i=1}^{a_x}$ functions.

On the one hand, the $\{\alpha_i(s)\}_{i=1}^{a_x}$ functions for the first kind of artificial datasets are built taking into account a similarity $\delta$ function applied to the side information $s$ of the target and a set of other side information descriptions $\{\mu^k\}_{k=1}^d$ . The idea is that the generated target value $y$ will be defined in terms of similarity values. Hence, the $\{\alpha_i(s)\}_{i=1}^{a_x}$ functions are defined as a weighted average of similarities:
\[
\alpha_i(s)=\frac{\sum_{k=1}^{d} \left(\tau_{i,k}\cdot \delta(s,\mu^k) \right)}{\sum_{k=1}^{d}\delta(s,\mu^k)} 
\]
The coefficients $\{\tau_{i,k}\}_{i=1, k=1}^{a_x, d}$ and the side information descriptions $\{\mu^{k}\}_{k=1}^{d}$ were drawn using the same distribution as for the feature values of $X$ and $S$, which is a uniform distribution in the range of $(-2;-1] \cup [+1;+2)$. The similarity function $\delta$ has been randomly chosen to be either the Manhattan (L1 norm) or the Euclidean (L2 norm) in equal shares in order to avoid bias.

On the other hand, the $\{\alpha_i(s)\}_{i=1}^{a_x}$ functions for the second kind of artificial datasets are built taking the most general structure that provides a linear dependence of side information. That is:

\[
\alpha_i(s)=\sum_{j=1}^{a_s}(\gamma_{i,j}\cdot s_j)+\beta_i
\]

The coefficients $\beta$, $\{\beta_i\}_{i=1}^{a_x}$ and $\{\gamma_{i,j}\}_{i=1, j=1}^{a_x, a_s}$ were drawn using the same distribution as for the feature values of $X$ and $S$, the coefficients $\{\tau_{i,k}\}_{i=1, k=1}^{a_x, d}$ and the side information descriptions $\{\mu^{k}\}_{k=1}^{d}$, which is a uniform distribution in the range of $(-2;-1] \cup [+1;+2)$.

Without loss of generality, the number of instances and features were fixed to 5000 and 50, respectively. However, due to the focus of the paper, the number of targets and features that conform the side information were varied to obtain different datasets. The number of targets considered were ${5, 10, 50}$ and ${100}$, and the number of side information features were ${5, 15}$ and ${25}$, in order to cover a range both below and above the real datasets. The number of the other side information descriptions $d$ for the first kind of datasets was fixed as $5$ in order to allow the dataset with $5$ targets to include enough information to be learned. Table \ref{tab:artificialdatasets} displays the main properties of these datasets. The datasets are named $\text{S}^{k,l}$ (first kind of artificial datasets) and $\text{R}^{k,l}$ (second kind of artificial datasets) where $k$ and $l$ respectively indicate the number of targets and the size of the side information.

\begin{table}[H]
	\begin{center}
		\begin{tabular}{ l | r | c  || l | r | c } 
			\hline
 \multirow{2}{*}{dataset}&\multirow{2}{*}{targets}&side&\multirow{2}{*}{dataset}&\multirow{2}{*}{targets}&side\\
    &&information&&& information\\ 
			\hline
			$\text{S}^{5,5}/\text{R}^{5,5}$ & 5 & 5 &$\text{S}^{50,15}/\text{R}^{50,15}$ & 50 & 15 \\
			$\text{S}^{10,5}/\text{R}^{10,5}$ &  10 & 5 &$\text{S}^{100,15}/\text{R}^{100,15}$ &  100 & 15\\
			$\text{S}^{50,5}/\text{R}^{50,5}$ & 50 & 5 &$\text{S}^{5,25}/\text{R}^{5,25}$ & 5 & 25 \\
			$\text{S}^{100,5}/\text{R}^{100,5}$ & 100 & 5 &$\text{S}^{10,25}/\text{R}^{10,25}$ & 10 & 25 \\
			$\text{S}^{5,15}/\text{R}^{5,15}$ &  5 & 15 &$\text{S}^{50,25}/\text{R}^{50,25}$ &  50 & 25 \\
			$\text{S}^{10,15}/\text{R}^{10,15}$ & 10 & 15 &$\text{S}^{100,25}/\text{R}^{100,25}$ & 100 & 25 \\
		        \hline
		\end{tabular}
		\caption{Number of targets and side information size of the artificial datasets}
		\label{tab:artificialdatasets}
	\end{center}
\end{table}

\subsubsection{Real datasets: Communities and Crime dataset and Air pollution dataset}
Two real datasets were selected in the experiments to check the two proposed methods and to compare them with the baseline; namely, the Communities and Crime dataset and the Air pollution dataset. Table \ref{tab:realdatasets} shows the main properties of both datasets. The descriptions of both datasets are detailed below.
\begin{table}[H]
	\begin{center}
		\begin{tabular}{ l | r | r | r | c } 
			\hline
                         \multirow{2}{*}{dataset}&\multirow{2}{*}{instances}&\multirow{2}{*}{features}&\multirow{2}{*}{targets}&side\\
    &&&&information\\ 
			\hline
			Communities and Crime & 1420 & 101 & 13 & 15 \\
			Air pollution & 41325 & 12 & 5 & 16 \\ 
		\hline
		\end{tabular}
		\caption{Properties of the real datasets:  Communities and Crime dataset and Air pollution dataset}
		\label{tab:realdatasets}
	\end{center}
\end{table}

\paragraph{Communities and Crime dataset}
The Communities and Crime dataset of the UCI Machine Learning Repository\footnote{https://archive.ics.uci.edu/ml/index.php} combines socioeconomic data from the 1990 US Census, law enforcement data from the 1990 US LEMAS survey, and crime data from the 1995 FBI UCR \cite{crimedataset}. The ratio of violent crimes per capita is obtained from the population and the sum of crime considered violent in various counties of the United States of America: murder, rape, robbery, and assault. The original version of this dataset is not suitable for zero-shot regression, so it had to be adapted. The instances are counties whose features are socioeconomic characteristics. The counties form states, that will become the targets, for which socioeconomic features are also taken as side information. Hence, the goal is to predict the crime ratio for a county belonging to a new state. 

\paragraph{Air pollution dataset}
The air pollution dataset was the motivation for this research and includes hourly averages of air pollution of the Principality of Asturias, Spain, from 2010 to 2018. This dataset corresponds to 11 types of pollutants from 18 pollution and meteorological stations. However, the largest common set of pollutants and stations included 4 pollutants ($\text{NO}_{2}$, TSP, NO, $\text{SO}_{2}$) from 5 stations, since not all the stations register the same air pollutants.

\begin{figure}[h]
	\centering
	\includegraphics[width=10cm]{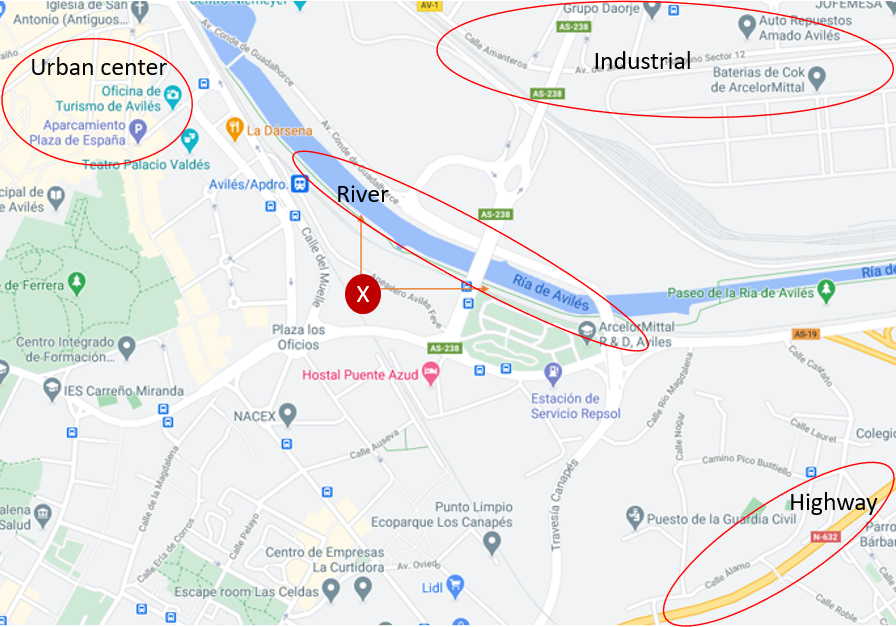}
	\caption{Example of a location with different surroundings elements. In relation to point $X$, the urban center is located to the north and west, a highway is located to the south and east, an industrial area is located to the north and east and finally, a river or sea is located to the north and east.}
	\label{fig:directionsMap}
	\centering
\end{figure}

In addition to pollutant measurements, meteorological data is taken into account, such as wind direction and speed, season, temperature, humidity, pressure or precipitation. External data is also available concerning certain properties of the surroundings of the stations, such as if there is an urban center, a highway, an industrial area, the sea or a river nearby. These are represented by indicating their directions (north, south, east and/or west). Hence, a total of $16$ features describe the stations in the form of side information. Figure \ref{fig:directionsMap} displays an example with a location that has an urban area in the northwest, a highway in the southeast and an industrial area and a river towards the northeast. 


The aim is to predict the amount of a certain pollutant in a location where a new station may be installed, for which meteorological data has not been collected. 

\subsection{Parameter settings} \label{settings}
Among the baseline method, SR and MPLC, only the SR has a parameter, namely, the distance for the relationship. In this sense, both the Manhattan (L1 norm) and the Euclidean (L2 norm) distances were varied for the relationship between observed and unobserved targets. They were taken as a preliminary approximation, since they are typical measures adopted by the k-nearest neighbor method.

\begin{figure}[h]
	\centering
	\includegraphics[width=10cm, trim={0 3cm 0 3cm}, clip]{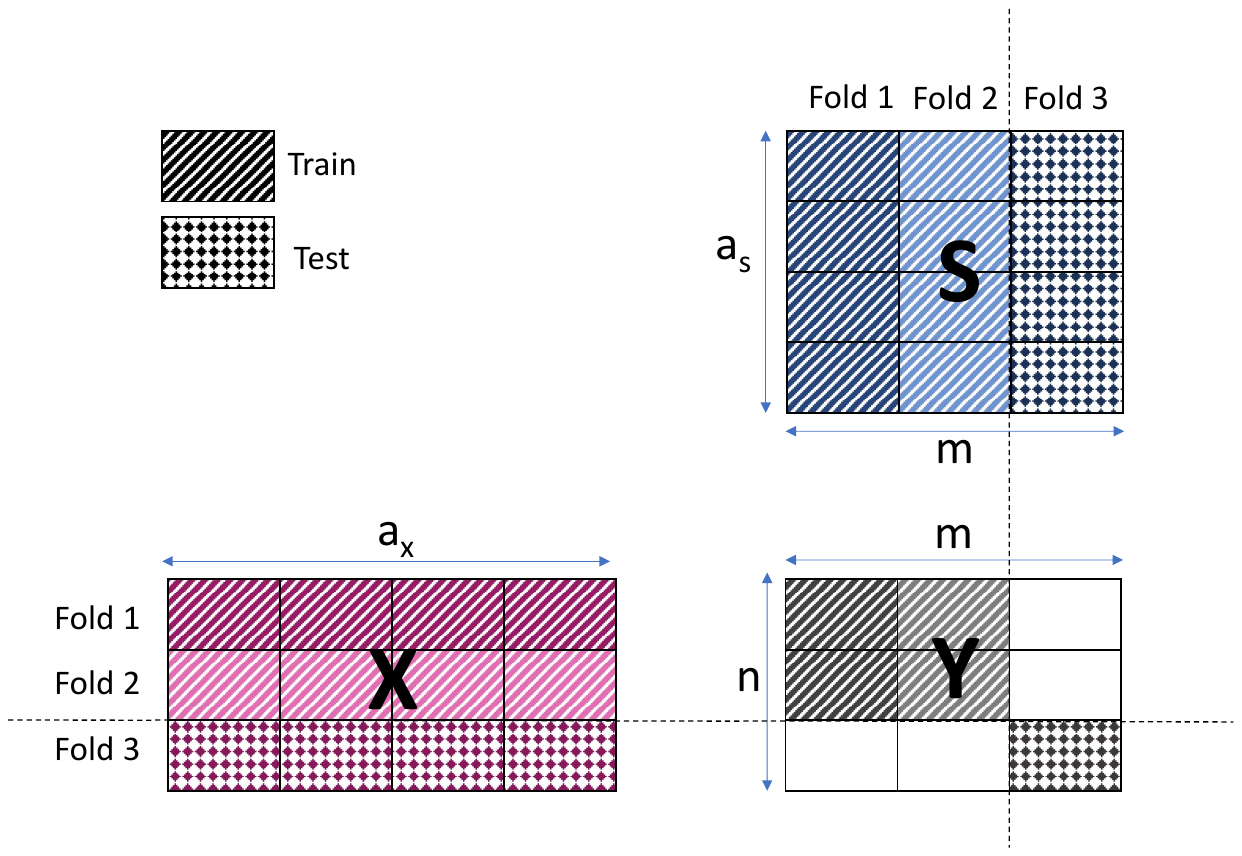}
	\caption{The matrices X, Y and S will be cut in a consistent way, but not all of the data is going to be used in each fold (white without texture).}
	\label{fig:crossValidation}
	\centering
\end{figure}

All the approaches for the experiments were implemented in Python using \textit{Scikit Learn Library} \footnote{\url{https://scikit-learn.org/stable/}}.  All the code is available in \href{https://github.com/UO231492/MPLC}{\texttt{GitHub repository}.} The same regressor was used as a base learner to induce the models of the observed targets and, in the case of the MPLC, to induce the models for learning the parameters of an unobserved target. Two different regressors were checked, namely, i) Ridge Regression (Ridge) \footnote{\url{https://scikit-learn.org/stable/modules/generated/sklearn.linear_model.Ridge.html}} and ii) Linear Support Vector Regression (LSVR) \footnote{\url{https://scikit-learn.org/stable/modules/generated/sklearn.svm.LinearSVR.html}}. Ridge solves a regression model where the loss function is the linear least squares function and regularization is given by the L2 norm. LSVR solves a convex quadratic optimization problem with linear constraints, similar to Support Vector Regression (SVR), which uses a linear kernel. The $\alpha$ parameter of Ridge and $c$ parameter of LSVR were optimized through a grid search using the mean squared error as the evaluation function and were estimated via a 3-fold cross validation. The values of $\alpha$ checked in the grid search for Ridge were $\{e^{-3},e^{-2},\dots, e^{2}, e^{3}\}$ and the values of $c$ checked in the grid search for LSVR were $\{10^{-3},10^{-2},\dots, 10^{2}, 10^{3}\}$. The score for the results was the relative mean squared error computed following a 3-fold cross validation with 3 repetitions. This cross validation, which has side information present, differs slightly from that of classic machine learning. In this case, the split in folders requires us to take into account instances and targets, and it must be consistent. Figure \ref{fig:crossValidation} illustrates an example where the $X$, $Y$ and $S$ matrices are split in 3 folds (2 for training and 1 for testing). The striped cells are for training, whereas the dotted cells are for testing. In the case of $Y$, the blank cells in the fold that play the role of testing are ignored in order to make the split consistent. These cells correspond to the values of the targets: i) for the observed (training) targets and unobserved (testing) instances (the blank cells of $Y$ on the bottom), and ii) for the unobserved (testing) targets and observed (training) instances (the blank cells of $Y$ on the right).

\subsection{Result analysis and discussion}
\label{results}
This section analyzes and discusses the results obtained from the experiments carried out over the above-described datasets, using the baseline and the two proposed methods: SR and MPLC. Firstly, the discussion is centered on the artificial datasets and afterwards it focuses on the real datasets about crime and pollution. The tables will display the relative mean squared error \cite{morgan2005dm}, which is the mean squared error normalized by the mean squared error of the default predictor (the mean of the predictions). The Friedman-Nemenyi test \cite{demvsar2006statistical,pacifico2018plant} was carried out. This test consists of a two-step comparison for each measure. The first step is a Friedman test \cite{friedman1937use,friedman1940comparison}, which rejects the null hypothesis that states that not all learners perform equally. The second step is the post-hoc Nemenyi test \cite{nemenyi1962distribution}. In the tables, the ranks of each dataset are shown in brackets. In case of ties, average ranks are indicated. The average ranks over all the datasets are calculated and presented in the last row of each table.

\begin{table}[p]
	\begin{tabular}{lrrrr}
		\hline
		
		\multirow{2}{*}{\textbf{Ridge}} &\multicolumn{1}{c}{\multirow{2}{*}{baseline}}&\multicolumn{1}{c}{SR}&\multicolumn{1}{c}{SR}&\multicolumn{1}{c}{\multirow{2}{*}{MPLC}}\\
		&&(Euclidean)&(Manhattan)&\\
		\hline
$\text{S}^{5,5}$&\textbf{92.17}(1)&97.82(3)&95.97(2)&111.29(4)\\ 
$\text{S}^{10,5}$&56.00(4)&50.51(3)&50.15(2)&\textbf{44.30}(1)\\ 
$\text{S}^{50,5}$&58.86(4)&50.02(3)&49.20(2)&\textbf{26.55}(1)\\ 
$\text{S}^{100,5}$&45.42(4)&38.11(3)&37.53(2)&\textbf{15.74}(1)\\ 
$\text{S}^{5,15}$&3.64(3)&3.46(2)&\textbf{3.28}(1)&94.73(4)\\ 
$\text{S}^{10,15}$&2.19(3)&\textbf{2.13}(1)&2.14(2)&75.32(4)\\ 
$\text{S}^{50,15}$&2.69(4)&2.43(3)&2.39(2)&\textbf{0.14}(1)\\ 
$\text{S}^{100,15}$&1.94(4)&1.84(3)&1.83(2)&\textbf{0.14}(1)\\ 
$\text{S}^{5,25}$&2.86(3)&2.83(2)&\textbf{2.79}(1)&81.28(4)\\ 
$\text{S}^{10,25}$&\textbf{0.83}(2)&\textbf{0.83}(2)&\textbf{0.83}(2)&98.43(4)\\ 
$\text{S}^{50,25}$&1.43(4)&1.36(3)&1.35(2)&\textbf{0.18}(1)\\ 
$\text{S}^{100,25}$&1.18(4)&1.14(3)&1.13(2)&\textbf{0.06}(1)\\ 
		\hline
		Avg. Rank&\ \ \  \ (3.3)&\ \ \  \ (2.6)&\ \ \  \ (\textbf{1.8})&\ \ \  \ (2.3)\\ \hline
		\multirow{2}{*}{\textbf{LSVR}} &\multicolumn{1}{c}{\multirow{2}{*}{baseline}}&\multicolumn{1}{c}{SR}&\multicolumn{1}{c}{SR}&\multicolumn{1}{c}{\multirow{2}{*}{MPLC}}\\
		&&(Euclidean)&(Manhattan)&\\
		\hline	
$\text{S}^{5,5}$&\textbf{91.95}(1)&96.83(3)&94.96(2)&121.80(4)\\ 
$\text{S}^{10,5}$&55.86(3)&48.76(2)&\textbf{48.16}(1)&60.13(4)\\ 
$\text{S}^{50,5}$&58.84(4)&49.73(3)&48.93(2)&\textbf{26.55}(1)\\ 
$\text{S}^{100,5}$&45.42(4)&38.11(3)&37.53(2)&\textbf{15.74}(1)\\ 
$\text{S}^{5,15}$&3.73(3)&3.38(2)&\textbf{3.21}(1)&99.83(4)\\ 
$\text{S}^{10,15}$&\textbf{2.22}(1)&2.39(2)&2.40(3)&77.73(4)\\ 
$\text{S}^{50,15}$&2.69(4)&2.43(3)&2.39(2)&\textbf{0.14}(1)\\ 
$\text{S}^{100,15}$&1.94(4)&1.84(3)&1.83(2)&\textbf{0.14}(1)\\ 
$\text{S}^{5,25}$&\textbf{3.26}(1)&3.46(3)&3.40(2)&93.85(4)\\ 
$\text{S}^{10,25}$&\textbf{0.90}(1)&1.48(3)&1.47(2)&98.89(4)\\ 
$\text{S}^{50,25}$&1.43(4)&1.36(3)&1.35(2)&\textbf{0.18}(1)\\ 
$\text{S}^{100,25}$&1.19(4)&1.13(3)&1.12(2)&\textbf{0.06}(1)\\ 
		\hline
		Avg. Rank&\ \ \  \ (2.8)&\ \ \  \ (2.8)&\ \ \  \ (\textbf{1.9})&\ \ \  \ (2.5)\\ \hline
	\end{tabular}
	\caption{Relative mean square error and Friedman-Nemenyi test for S artificial datasets using baseline, SR (Euclidean), SR (Manhattan) and MPLC}
	\label{table:FriedmanArtificialKnnD0}
\end{table}

\begin{table}[p]
	\begin{tabular}{lrrrr}
		\hline

		\multirow{2}{*}{\textbf{Ridge}} &\multicolumn{1}{c}{\multirow{2}{*}{baseline}}&\multicolumn{1}{c}{SR}&\multicolumn{1}{c}{SR}&\multicolumn{1}{c}{\multirow{2}{*}{MPLC}}\\
		&&(Euclidean)&(Manhattan)&\\
		\hline
		$\text{R}^{5,5}$&108.68(4)&96.23(3)&95.49(2)&\textbf{61.55}(1)\\ 
		$\text{R}^{10,5}$&121.34(4)&93.69(3)&86.84(2)&\textbf{1.45}(1)\\ 
		$\text{R}^{50,5}$&102.51(4)&69.05(3)&63.44(2)&\textbf{0.00}(1)\\ 
		$\text{R}^{100,5}$&102.38(4)&68.50(3)&62.97(2)&\textbf{0.00}(1)\\ 
		$\text{R}^{5,15}$&150.44(4)&148.66(3)&146.43(2)&\textbf{83.47}(1)\\ 
		$\text{R}^{10,15}$&114.71(4)&111.58(3)&108.60(2)&\textbf{72.30}(1)\\ 
		$\text{R}^{50,15}$&103.20(4)&96.09(3)&92.46(2)&\textbf{0.00}(1)\\ 
		$\text{R}^{100,15}$&101.64(4)&94.25(3)&90.13(2)&\textbf{0.00}(1)\\ 
		$\text{R}^{5,25}$&136.30(4)&132.95(3)&129.55(2)&\textbf{83.71}(1)\\ 
		$\text{R}^{10,25}$&119.76(4)&114.60(3)&111.88(2)&\textbf{65.06}(1)\\ 
		$\text{R}^{50,25}$&102.32(4)&98.24(3)&96.00(2)&\textbf{0.00}(1)\\ 
		$\text{R}^{100,25}$&102.22(4)&97.77(3)&95.21(2)&\textbf{0.00}(1)\\ 
		\hline
		Avg. Rank&\ \ \  \ (4.0)&\ \ \  \ (3.0)&\ \ \  \ (2.0)&\ \ \  \ \textbf{(1.0)}\\  \hline
		\multirow{2}{*}{\textbf{LSVR}} &\multicolumn{1}{c}{\multirow{2}{*}{baseline}}&\multicolumn{1}{c}{SR}&\multicolumn{1}{c}{SR}&\multicolumn{1}{c}{\multirow{2}{*}{MPLC}}\\
		&&(Euclidean)&(Manhattan)&\\
		\hline
		$\text{R}^{5,5}$&103.14(4)&88.28(3)&87.79(2)&\textbf{73.53}(1)\\ 
		$\text{R}^{10,5}$&119.56(4)&94.01(3)&87.96(2)&\textbf{2.10}(1)\\ 
		$\text{R}^{50,5}$&102.49(4)&69.05(3)&63.44(2)&\textbf{0.00}(1)\\ 
		$\text{R}^{100,5}$&102.36(4)&68.50(3)&62.97(2)&\textbf{0.00}(1)\\ 
		$\text{R}^{5,15}$&142.28(4)&130.19(3)&128.48(2)&\textbf{87.81}(1)\\ 
		$\text{R}^{10,15}$&112.80(4)&104.77(3)&102.64(2)&\textbf{72.07}(1)\\ 
		$\text{R}^{50,15}$&103.12(4)&96.09(3)&92.46(2)&\textbf{0.00}(1)\\ 
		$\text{R}^{100,15}$&101.63(4)&94.25(3)&90.13(2)&\textbf{0.00}(1)\\ 
		$\text{R}^{5,25}$&129.51(4)&118.60(3)&116.24(2)&\textbf{85.12}(1)\\ 
		$\text{R}^{10,25}$&117.85(4)&108.00(3)&106.01(2)&\textbf{66.81}(1)\\ 
		$\text{R}^{50,25}$&102.26(4)&98.30(3)&96.14(2)&\textbf{0.00}(1)\\ 
		$\text{R}^{100,25}$&102.21(4)&97.82(3)&95.21(2)&\textbf{0.00}(1)\\
		\hline
		Avg. Rank&\ \ \  \ (4.0)&\ \ \  \ (3.0)&\ \ \  \ (2.0)&\ \ \  \ \textbf{(1.0)}\\  \hline
	\end{tabular}
	\caption{Relative mean square error and Friedman-Nemenyi test for R artificial datasets using baseline, SR (Euclidean), SR (Manhattan) and MPLC}
	\label{table:FriedmanArtificialLinear}
\end{table}

\subsubsection{Artificial datasets}
Tables \ref{table:FriedmanArtificialKnnD0} and \ref{table:FriedmanArtificialLinear} display the relative mean square error and Friedman-Nemenyi test when baseline, SR (Euclidean), SR (Manhattan) and MPLC methods are applied to the artificial datasets. The results experimentally confirm the theoretical hypothesis (also shown through an illustrative toy example) about the loss of knowledge that takes place when information about the observed target models is not used for learning the unobserved target models. Effectively, the baseline method achieves the worst average rank, regardless of the regressor used for estimating the error (Ridge or LSVR) and the artificial datasets (S and R datasets). In S artificial datasets, and taking into account the average ranks, SR (Manhattan) outperforms the rest of the methods, followed by MPLC and then by SR (Euclidean). However, focusing on the particular ranks of MPLC, one can clearly see that MPLC comes first for half of the S artificial datasets, specifically those with the highest number of targets (50 and 100), regardless of the number of features of the side information. MPLC does not outperform SR (Manhanttan) due to the fourth place that it yields when the number of targets is low ($5$ and $10$).  In R artificial datasets, MPLC clearly obtains the best scores, regardless of the regressor used for estimating the error (Ridge or LSVR). Indeed in some cases the error of MPLC is practically $0$.

Since both Table \ref{table:FriedmanArtificialKnnD0} and Table \ref{table:FriedmanArtificialLinear} shows the relative error, only MPLC are unable to achieve the media system in one S artificial dataset ($\text{S}^{5,5}$), which corresponds to the S dataset with the lowest number of targets and the lowest size of side information. However, in the case of the R artificial datasets, the baseline approach is the method that is unable to outperform the media system.  This is also the case for both versions of SR in the instances of $\text{R}^{5,15}$, $\text{R}^{10,15}$, $\text{R}^{5,25}$ and $\text{R}^{10,25}$. Meaning that SR does not work well for datasets with few targets and a high number of side information features when the datasets are not generated using similarities. Focusing on the similarity measures of SR, Manhattan distance seems to perform slightly better than Euclidean distance for all the datasets, whatever S and R artificial datasets.

In case of the S artificial datasets, SR benefits when the side information size is sufficiently high (from $15$), whatever the number of targets, since the relative error of both the SR versions falls considerably to values below $4\%$. However, in the case of the R artificial datasets, SR benefits most when the datasets are characterized as having the highest number of targets and fewer side information features ($\text{R}^{50,5}$ and $\text{R}^{100,5}$).

MPLC seems to benefit from having a high number of observed targets ($50$ and $100$), since the relative error falls drastically under these situations and the method comes out in the first place, whatever the S or R artificial datasets. In case of the S artificial datasets, the relative error falls below $1\%$ when the side information size is sufficiently high (from $15$). In the case of the R artificial datasets,  the error equals $0\%$, no matter how large the feature side information is. This is consistent with having enough instances to learn the parameters, since the number of targets conditions the number of instances in the second learning phase. 

In the case of the R artificial datasets, it should be noted that the methods, when ranked by performance, remain in the same order regardless of the dataset chosen. However, in the case of the S artificial datasets, the baseline method and MPLC seem to distribute the first and fourth places, whereas SR (Euclidean) and SR (Manhattan) seem to distribute the second and third places.

A Wilcoxon test \cite{wilcoxon1992individual} was made using the ranks of each pair of systems. Since the comparison is between pairs of methods, the ranks take values $1$ or $2$. The size of the sample for the test is $24=12\cdot 2$, which corresponds with the $12$ artificial datasets and the $2$ base learners. In the case of the S artificial dataset, SR (Manhattan) performs better than SR (Euclidean) with a $p-$value equal to $0.0067$ in the case of Ridge and $0.0039$ in the case of LSVR. Also, in the case of the S artificial dataset, both SR (Manhattan) and SR (Euclidean) outperforms the baseline method with a $p-$value equal to $0.0067$. In the case of the R artificial datasets, the $p-$value is the same for each pair of systems and equal to $0.0005$. This is because MPLC always performs better than SR (Manhattan), SR (Manhattan) always performs better than SR (Euclidean) and in turn SR (Euclidean) always outperforms baseline. This $p-$value far exceeds even the risk of $1\%$, and therefore the differences in performance between pairs of methods are statistically significant.

\begin{table}[h]
	\begin{tabular}{lrrrr}
		\hline
		\multirow{2}{*}{\textbf{Ridge}} &\multicolumn{1}{c}{\multirow{2}{*}{baseline}}&\multicolumn{1}{c}{SR}&\multicolumn{1}{c}{SR}&\multicolumn{1}{c}{\multirow{2}{*}{MPLC}}\\
		&&(Euclidean)&(Manhattan)&\\
		\hline
		Crime&\textbf{27.56}(1)&36.52(3)&34.20(2)&51.65(4)\\
		$\text{NO}_{2}$&86.32(4)&84.19(3)&80.24(2)&\textbf{78.67}(1)\\ 
		PST&92.40(3)&92.55(4)&90.86(2)&\textbf{87.65}(1)\\ 
		NO&101.01(4)&100.73(2)&100.75(3)&\textbf{89.31}(1)\\ 
		$\text{SO}_{2}$&95.14(2)&96.26(4)&95.19(3)&\textbf{93.63}(1)\\
		 \hline 
		Avg. Rank&\ \ \  \ (2.8)&\ \ \  \ (3.2)&\ \ \  \ (2.4) \ \ \  \ &\textbf{(1.6)}\\ \hline
	
		\multirow{2}{*}{\textbf{LSVR}} &\multicolumn{1}{c}{\multirow{2}{*}{baseline}}&\multicolumn{1}{c}{SR}&\multicolumn{1}{c}{SR}&\multicolumn{1}{c}{\multirow{2}{*}{MPLC}}\\
		&&(Euclidean)&(Manhattan)&\\
		\hline
		Crime&33.36(2)&33.43(3)&35.86(4)&\textbf{25.40}(1)\\ 
		$\text{NO}_{2}$&86.11(4)&81.07(3)&76.52(2)&\textbf{76.43}(1)\\ 
		PST&90.86(3)&91.17(4)&88.95(2)&\textbf{87.15}(1)\\ 
		NO&96.20(4)&93.84(3)&\textbf{90.28}(1)&91.68(2)\\ 
		$\text{SO}_{2}$&95.01(3)&95.53(4)&\textbf{92.19}(1)&93.49(2)\\ 
		\hline
		Avg. Rank&\ \ \  \ (3.2)&\ \ \  \ (3.4)&\ \ \  \ (2.0)\ \ \  \ &\textbf{(1.4)}\\ \hline
	\end{tabular}
	\caption{Relative mean square error and Friedman-Nemenyi test for the Communities and Crime and air pollution datasets using baseline, SR (Euclidean), SR (Manhattan) and MPLC}
	\label{table:FriedmanReal}
\end{table}

\subsubsection{Real datasets}
This section discusses the experiment results over the Communities and Crime and the air pollution datasets. Table \ref{table:FriedmanReal} displays the relative mean square error and Friedman-Nemenyi test when baseline, SR (Euclidean), SR (Manhattan) and MPLC methods are applied over these real datasets. Our conclusions remain similar in the experiments carried out over these datasets, namely, i) MPLC shows the best performance and ii) Between the two versions of SR, Manhattan distance seems to work better and comes close to MPLC performance. Two particulars stand out in the experiments and deserve comment. The first one is the high performance that the baseline method attains for the Communities and Crime dataset when Ridge regression is used. The other one is that SR (Manhattan) slightly outperforms MPLC for two of the four pollutants. Focusing on the relative errors, it seems that NO pollutant is quite hard to learn, since only the MPLC method manages to overcome the media system using Ridge regression. However, $\text{NO}_{2}$ is the easiest to learn when compared to the rest of pollutants. The $p-$values of the Wilcoxon test are displayed in Table \ref{table:wilconxonreal}. The size of the sample in this case is $10=5\cdot 2$, because of the $5$ real datasets and the $2$ base learners. 
These $p-$values indicate that there are no significant differences between baseline and SR (Euclidean), or between SR (Manhattan) and MPLC. However, there are significant differences between MPLC and baseline, as well as between MPLC and SR (Euclidean), at a risk of $1\%$. There are also significant differences between both versions of SR, Euclidean and Manhattan, although in this case at a risk of $5\%$. In consequence, MPLC shows itself to be superior to the rest of the methods, but SR (Manhattan) performs quite similarly.

\begin{table}[H]
	\begin{tabular}{lcccc}
		\hline
&\multicolumn{1}{c}{SR}&\multicolumn{1}{c}{SR}&\multicolumn{1}{c}{\multirow{2}{*}{MPLC}}\\
&(Euclidean)&(Manhattan)&\\
  \hline
  baseline  & 0.52& 0.20&0.01\\
  SR (Euclidean) & & 0.05&0.01\\
  SR (Manhattan) & & &0.20\\
  \hline
	\end{tabular}
	\caption{$p$-values of the Wilcoxon test for the communities and crime, and air pollution datasets}
	\label{table:wilconxonreal}
\end{table}      
	
\section{Conclusions and future work} \label{conclusions}
This paper proposes two approaches to cope with regression zero-shot tasks. To the best of our knowledge, zero-shot has been mainly focused on computer vision, speech or character recognition classification tasks. The interest in providing methods for zero-shot regression in this paper was spurred by the need to predict air pollution in locations where weather conditions have not been registered before, but where environmental information about the location is available. This information can be considered side information, since it is neither features nor targets, it is known beforehand, is independent of possible unobserved weather conditions, and remains constant for the same location. It is a potential source of information susceptible to being exploited that makes the task suitable for a zero-shot regression environment. 

One of the contributions of this paper is a similarity based relationship zero-shot method for regression that combines the feature (weather conditions) models for the observed targets (meteorological stations) in order to obtain models for the unobserved targets (potential future locations of new meteorological stations). However, this method has the drawback of losing potential knowledge of the feature models, since it just performs an aggregation procedure, limiting the power of generalization to locations of future meteorological stations. The second contribution of the paper is a parameter learning model based correspondence zero-shot method for regression that overcomes this limitation by attempting not to aggregate the predictions of the feature models, but extracting knowledge directly from the models, including the parameters within a learning procedure. This learning procedure is what provides the generalization power in order to obtain predictions for unobserved stations from weather conditions. Both proposals are compared to a simple baseline method, consisting in taking both weather conditions (features) and environmental information (side information) as common features. Several artificial data sets are generated so that a more detailed comparison can be performed, by means of varying the number of targets and the size of the side information. The comparison is also carried out using the Communities and Crime dataset of the UCI Machine Learning Repository, in addition to the Air pollution dataset.

The conclusion is that both the parameter learning model based correspondence method and the similarity based relationship method with Manhattan distance outperform both similarity based relationship method with Euclidean distance and the baseline approaches, the improvement in performance being statistically significant for Communities and Crime dataset of the UCI Machine Learning Repository and the Air pollution dataset. The parameter learning model based correspondence method outperforms the similarity based relationship method with Manhattan distance for some artificial datasets, whereas the latter obtains better performance than the former for other artificial datasets. As far as the variations in the number of observed targets and the size of the side information are concerned, a sufficient number of observed targets is crucial in order to obtain promising parameter learning, whilst; also, having fewer side information features brings a better performance. Whichever the case, these contributions enrich the zero-shot framework in relation to regression tasks.

When it comes to possible future work, several lines can be opened. One potentially interesting future line of research would be adapting the strategies proposed in this paper to neural networks, and particularly to deep learning. This adaptation lies outside of the scope of this paper, since such an adaptation is no trivial task, given the large amount of hyperparameters present in neural networks. Another future line may be developed if instead of limiting the task to regression, we extended the procedure to multiregressions. This extension may improve the performance of the predictions, since possible information about correlations among the targets could be exploited. In the context of air pollution, this means relating different pollutants. This practice is sensible, since the presence of certain pollutants may condition the presence or absence of others. Another line susceptible to being exploited might be to improve the parameter learning procedure, trying to simultaneously optimize both learning phases. With regard to the similarity based relationship zero-shot method for regression, it should be possible to explore and design new relationship functions in order to improve their performance. 
\section*{Acknowledgments}\label{sec:Acknowledgments}
This research has been partially supported by the Spanish Ministry of Science and Innovation through the grant PID2019-110742RB-I00.

\bibliographystyle{apalike}
\biboptions{numbers}
\bibliography{mybib}
\end{document}